\documentclass[accepted]{uai2023} 

\usepackage[american]{babel}

\usepackage{natbib} 
\usepackage{mathtools} 
\usepackage{booktabs} 
\usepackage{tikz} 

\usepackage{graphicx}
\usepackage{booktabs} 
\usepackage{subfig}
\usepackage{wrapfig}
\usepackage{algorithm}
\usepackage{algorithmic}
\usepackage{amsmath}
\usepackage{amssymb}
\usepackage{mathtools}
\usepackage{amsthm}

\newcommand{\highlight}[1]{\colorbox{blue!10}{#1}}

\newcommand{\A}{\mathcal{A}}
\newcommand{\gG}{\mathcal{G}}
\newcommand{\gS}{\mathcal{S}}

\newcommand{\X}{\mathcal{X}}

\usepackage{xspace}
\newcommand*{\eg}{{\it e.g.}\@\xspace}
\newcommand*{\ie}{{\it i.e.}\@\xspace}

\title{Stochastic Generative Flow Networks}

\author[1,2]{\href{mailto:<penny.ling.pan@gmail.com>}{Ling Pan\thanks{Equal contribution.}}}
\author[1,2]{Dinghuai Zhang$^{*}$}
\author[1,2]{Moksh Jain}
\author[3]{Longbo Huang}
\author[1,2,4]{Yoshua Bengio}
\affil[1]{%
    Mila - Qu\'ebec AI Institute
}
\affil[2]{%
    Universit\'e de Montr\'eal
}
\affil[3]{%
    Tsinghua University
}
\affil[4]{%
    CIFAR AI Chair
}

\begin{document}
\maketitle

\begin{abstract}
Generative Flow Networks (or GFlowNets for short) are a family of probabilistic agents that learn to sample complex combinatorial structures through the lens of ``inference as control''. They have shown great potential in generating high-quality and diverse candidates from a given energy landscape. However, existing GFlowNets can be applied only to deterministic environments, and fail in more general tasks with stochastic dynamics, which can limit their applicability. To overcome this challenge, this paper introduces Stochastic GFlowNets, a new algorithm that extends GFlowNets to stochastic environments. 
By decomposing state transitions into two steps, Stochastic GFlowNets isolate environmental stochasticity and learn a dynamics model to capture it. Extensive experimental results demonstrate that Stochastic GFlowNets offer significant advantages over standard GFlowNets as well as MCMC- and RL-based approaches, on a variety of standard benchmarks with stochastic dynamics.
\end{abstract}

\section{Introduction}
Recently, Generative Flow Networks~\citep[GFlowNets;][]{bengio2021flow,bengio2021foundations} have been successfully applied to a wide variety of tasks, including molecule discovery~\citep{bengio2021flow,jain2022multi}, biological sequence design~\citep{jain2022biological}, and robust scheduling~\citep{robust-scheduling}.
GFlowNets learn policies to generate objects $x \in \mathcal{X}$ sequentially, and are related to Monte-Carlo Markov chain (MCMC) methods~\citep{metropolis1953equation,hastings1970monte,andrieu2003introduction}, generative models~\citep{goodfellow2014generative,ho2020denoising}, and amortized variational inference~\citep{kingma2013auto}. 
The sequential process of generating an object following a policy bears a close resemblance to reinforcement learning~\citep[RL;][]{sutton2018reinforcement}.
Contrary to the typical reward-maximizing policy in RL~\citep{,mnih2015human,lillicrap2015continuous,haarnoja2017reinforcement,fujimoto2018addressing,haarnoja2018soft}, GFlowNets aim to learn a stochastic policy for sampling composite objects $x$ with probability \emph{proportional to the reward function} $R(x)$.
This is desirable in many real-world tasks where the diversity of solutions is important, and we aim to sample a diverse set of high-reward candidates, including recommender systems~\citep{kunaver2017diversity}, drug discovery~\citep{bengio2021flow,jain2022biological}, and sampling causal models from a Bayesian posterior~\citep{deleu2022bayesian}, 
among others. 

\begin{figure}[!h]
\centering
\includegraphics[width=1.0\linewidth]{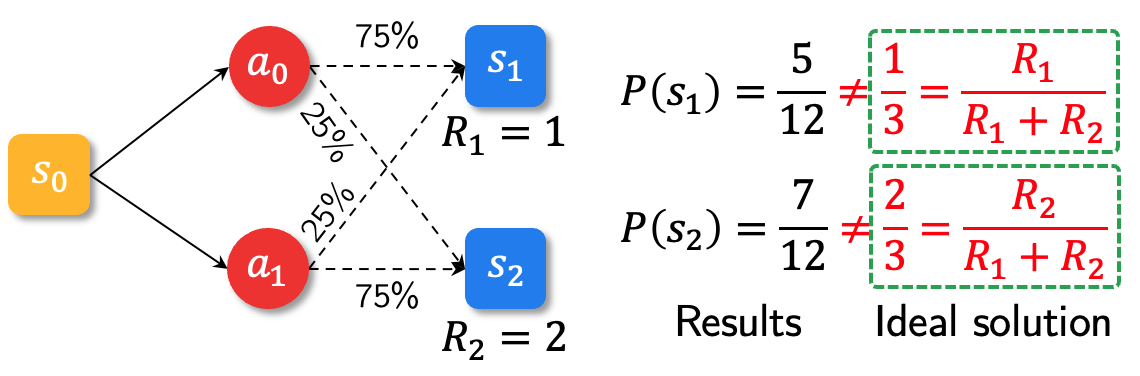}
\caption{An example illustrating the failure of existing GFlowNet approaches. (Left) Squares and circles denote states and actions, while solid and dotted arrows correspond to policy decisions and stochastic environment dynamics. The numbers above the dotted lines represent state transition probabilities, and the numbers below the blue squares (terminal states) denote the terminal reward. (Right) Results from existing GFlowNet approaches and the ideal solution.}
\label{fig:example}
\end{figure}

Existing work on GFlowNets~\citep{bengio2021flow,malkin2022trajectory,madan2022learning}, however, is limited to deterministic environments, where state transitions are deterministic, which may limit their applicability in the more general stochastic cases in practice. 
Figure~\ref{fig:example} illustrates an example with stochastic transition dynamics where existing GFlowNet approaches can fail.
Standard GFlowNet approaches will result in $P(s_1)=\frac{5}{12}$ and $P(s_2)=\frac{7}{12}$ when trained to completion (with $P(s)$ denoting the probability of sampling state $s$), which does not match the ideal case where $P(s_1)=\frac{1}{3}$ and $P(s_2)=\frac{2}{3}$. Therefore, the learned policy does not sample proportionally to the reward function in the presence of stochastic transition dynamics.
In practice, many tasks involve stochasticity in state transitions, which are more challenging to solve but are applicable to a wide variety of problems~\citep{antonoglou2021planning,yang2022dichotomy,paster2022you}.

To address this limitation, in this paper, we introduce a novel methodology, Stochastic GFlowNet, which is the first empirically effective approach for tackling environments with stochastic transition dynamics with GFlowNets.
Stochastic GFlowNet decomposes the state transitions based on the concept of \textit{afterstates}~\citep{sutton2018reinforcement,bengio2021foundations}.
Specifically, each stochastic state transition is decomposed into a deterministic step that transitions from the environment state $s$ to an intermediate state $(s, a)$ and a stochastic step that branches from the intermediate state $(s, a)$ to the next state $s'$.
We propose a practical way for training the dynamics model to capture the stochastic environment dynamics. The methodology is general and can be applied to different GFlowNet learning objectives.
The code is publicly available at
\url{https://github.com/ling-pan/Stochastic-GFN}.

In summary, the contribution of this work is as follows:
\begin{itemize}
    \item We propose a novel method, Stochastic GFlowNet, which is the first empirically effective approach extending GFlowNets to the more general stochastic environments based on \citet{bengio2021foundations}.
    \item We conduct extensive experiments on GFlowNet benchmark tasks augmented with stochastic transition dynamics, and validate the effectiveness of our approach in tackling stochastic environments. Results show that our method significantly outperforms existing baselines and scales well to the more complex and challenging biological sequence generation tasks. 
\end{itemize}

\section{Background}

\subsection{GFlowNet Preliminaries}
We denote a directed acyclic graph (DAG) by $\gG=(\gS, \A)$, with $\gS$ the set of vertices corresponding to the states and $\A\subseteq\gS\times\gS$ the set of edges, which corresponds to the set of actions. There is a unique \textit{initial state} $s_0\in\gS$ which has no parent state; on the other hand, we define all states without children to be \textit{terminal states}, whose set is denoted by $\X\subseteq\gS$.
A GFlowNet learns stochastic policy which aims to sample complete trajectories $\tau=(s_0\rightarrow s_1\rightarrow\dots\rightarrow s_n)$ where $s_n\in\X$ and $(s_i\rightarrow s_{i+1})\in{\A}, \forall i$ to sample terminal states. Each trajectory is assigned a non-negative \emph{flow} $F(\tau)$. A trajectory can be generated sequentially by sampling iteratively from the \textit{forward policy} $P_F(s_{t+1}|s_{t})$, which is a collection of distributions over the children at each state. Existing work on GFlowNets assumes a one-to-one correspondence between action and next state, making the definition of forward policy to be consistent to the notion of policy in general RL. Nonetheless, in this work, we relax this assumption and generalize GFlowNets to more flexible stochastic environments.
The objective of GFlowNet learning is to sample terminal states with probability proportional to a given non-negative reward function $R(x)$ for all $x\in\X$. 
This indicates that all the flows that end up with $x$ should sum up to $R(x)$, namely $\sum_{\tau\to x}F(\tau)=R(x),\forall x\in\X$, where $\tau\to x$ is a trajectory $\tau$ that ends in $x$ and the sum is thus over all complete trajectories that lead to terminal state $x\in\X$.
To formalize this, we first define the \textit{terminating probability} $P_T(x)$ to be the marginal likelihood of sampling trajectories terminating at a terminal state $x$:
\begin{align}
\label{eq:pt_pf_decomp}
    P_T(x) = \sum_{\tau\to x} P_F(\tau) = \sum_{\tau\to x}\prod_{i=1}^n P_F(s_i|s_{i-1}).
\end{align}
Therefore, the goal of GFlowNet learning is to obtain a policy such that $P_T(x)\propto R(x), \forall x\in\X$.

\subsection{Learning Objectives for GFlowNets}

In applicative tasks, practitioners need to design the GFlowNet modules (\eg, policies, flows) with parameterized neural networks, and further choose a training criterion to train these networks. In this subsection, we briefly summarize some learning criteria of GFlowNets.

\paragraph{Detailed balance (DB).}
By summing the flows $F(\tau)$ of all the trajectories $\tau$ going through a state $s$, we can define a state flow $F(s):=\sum_{\tau \ni s}F(\tau)$.
Such a function can be learned, together with forward and backward policy $P_F(s'|s), P_B(s|s')$, where $s'$ is a child state of $s$. The backward policy $P_B$, a collection of distributions over the parents of each state, is not part of the generative process, but serves as a tool for learning the forward policy $P_F$. The GFlowNet detailed balance (DB) constraint is defined as
\begin{align}
    F(s)P_F(s'|s) = F(s')P_B(s|s'),\ \forall (s\to s')\in\A.
\label{eq:db_cons}
\end{align}
It is also worth noting that at terminal states $x$, it pushes the flow at $x$ to match the terminal rewards $R(x)$.
In practice, we transform the DB constraint into a training objective by setting the loss function to be a squared difference between the logarithm of the left and right-hand sides~\cite{bengio2021flow} of Eq.~\eqref{eq:db_cons}.
If the optimization objective is perfectly minimized, it would make the above flow consistency constraint satisfied, thus making the forward policy $P_F$ sample proportionally to given reward values, as desired. 
It means that after training, the constraint is only approximately achieved (and in general it would be intractable to obtain an exact solution).

\paragraph{Trajectory balance (TB).}
In analogy to the forward decomposition of a complete trajectory in Eq.~\eqref{eq:pt_pf_decomp}, we could use $\prod_{i=1}^{n}P_B(s_{i-1}|s_i)$ to represent the trajectory backward probability. 
As an alternative to DB, \citet{malkin2022trajectory} propose the trajectory balance (TB) criterion which operates on complete trajctories, instead of state transitions, defined as follows
\begin{align}
    Z\prod_{i=1}^{n}P_F(s_{i}|s_{i-1}) = R(x)\prod_{i=1}^{n}P_B(s_{i-1}|s_i),
\end{align}
where $\tau= (s_0\to s_1\to\ldots\to s_n=x)$ is any complete trajectory and $Z$ is a learned scalar parameter, denoting the partition function of the reward distribution. Note that TB does not explicitly learn a flow function. 

For on-policy training, we can simply use trajectories sampled from the forward policy $P_F$ to evaluate the training loss and its gradient with respect to the parameters of the neural networks. The GFlowNet training objectives can be further improved with off-policy training, \ie, with trajectories sampled from a broader and more exploratory distribution than $P_F$~\citep{malkin2022gfnhvi}. A popular choice is using a tempered version of the current forward policy~\citep{zhang2022generative} or a mixture of the forward policy and a uniform random policy~\citep{bengio2021flow} that mimics $\epsilon$-greedy exploration in RL. 

\section{Stochastic GFlowNets} 

We now describe the Stochastic GFlowNet, a novel method that learns a model of the environment to capture the stochasticity of state transitions.
We first describe a key idea introduced by \citet{bengio2021foundations} to decompose the GFlowNet transitions as per Figure~\ref{fig:method}, and then introduce a new approach to learn the GFlowNet policy and the dynamics model.
We also discuss the applicability to different GFlowNet learning objectives and the resulting effects.

\subsection{Proposed Method} \label{sec:method}
Existing work on GFlowNets~\citep{bengio2021flow,malkin2022trajectory} typically makes the assumption that all transitions from a state $s_t$ to the next state $s_{t+1}$ within a trajectory are defined deterministically based on the selected action $a_t$ (and also there is only one action $a_t$ that can transition from $s_t$ to $s_{t+1}=T(s_t,a_t)$, with $T$ denoting the deterministic transition function). This applies to problems where the generative process for the objects is deterministic, which is appropriate when the actions are internal, \eg, choosing what to attend, what to imagine (such as solutions to problems), how to make inferences, etc.  
Yet, a number of real-world tasks are stochastic, either inherently or due to the environment complexity~\citep{antonoglou2021planning,paster2022you,yang2022dichotomy}.
In the more general stochastic environments, the action $a_t$ at $s_t$ can land in several possible next states. For instance, synthesizing proteins with oligo pools can result in the generation of variants of the specified protein~\citep{song2021large}.
\begin{figure}[!h]
\centering
\includegraphics[width=0.8\linewidth]{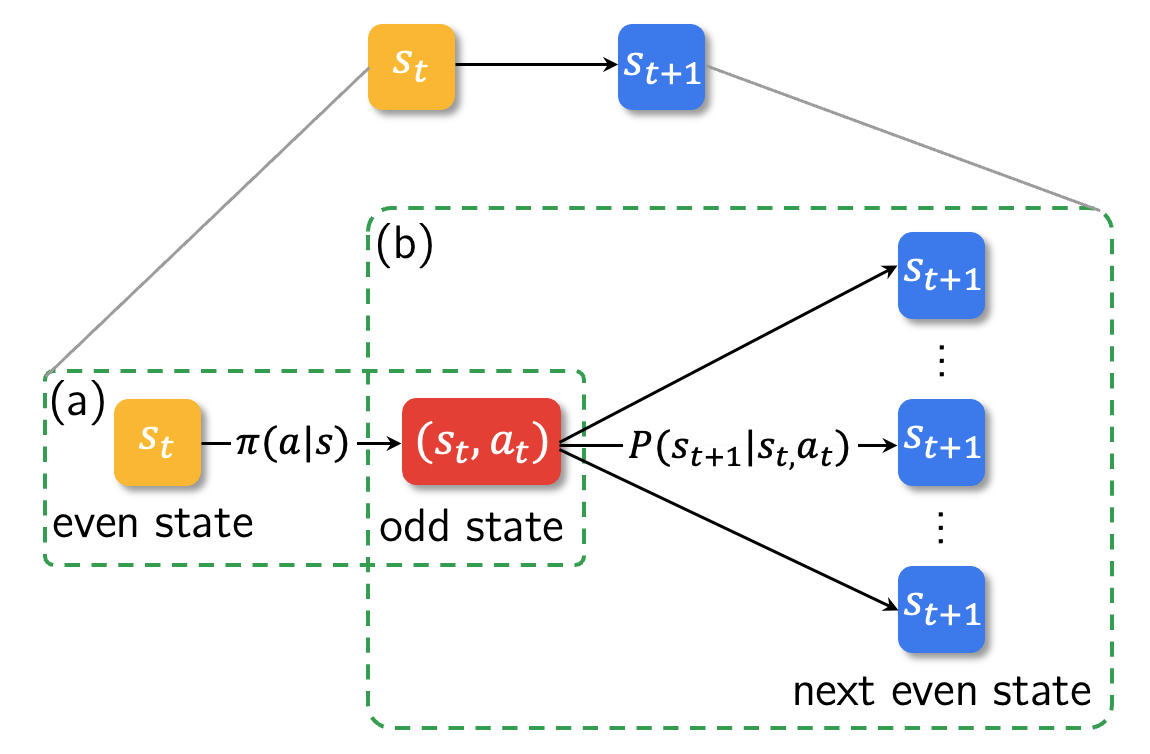}
\caption{We decompose traditional GFlowNet transitions (top) into two steps to facilitate the GFlowNet formalization: (a) stochastically choosing the action, (b) stochastically transitioning to a new state. We call the intermediate state after (1) an odd state (and starting and final points even states), as illustrated above.}
\label{fig:method}
\end{figure}

To cope with stochasticity in the transition dynamics, we decompose state transitions based on the concept of \textit{afterstates}~\citep{sutton2018reinforcement,bengio2021foundations}.
Specifically, for a transition from a state $s_t$ to the next state $s_{t+1}$, we decompose the transition in two steps.
First, as illustrated in part (a) in Figure~\ref{fig:method}, we sample an action $a_t$ based on a policy $\pi$ in the current state $s_t$ (called an even state), and transit deterministically to an intermediate state $(s_t, a_t)$, called an odd state.
The flow consistency constraint for detailed balance (DB) for even-to-add transitions is shown in Eq.~\eqref{eq:stoch_db1}, since the backward policy probability is $1$ here (we can only get to $(s_t,a_t)$ from $s_t$):
\begin{equation}
F(s_t) \pi(a_t | s_t) = F((s_t, a_t)).
\label{eq:stoch_db1}
\end{equation}
The odd state can be considered as a hypothetical state after we apply an action before the environment gets involved~\citep{antonoglou2021planning}. 
The environment dynamics then transform $(s_t,a_t)$ into to the next even state $s_{t+1}$ stochastically according to a distribution $P(\cdot|s_t,a_t)$, which is the state transition function.
This second step corresponds to part (b) in Figure~\ref{fig:method}, and the corresponding flow consistency constraint is
\begin{equation}
F((s_t, a_t))P(s_{t+1}|(s_t, a_t)) = F(s_{t+1}) \pi_B((s_t, a_t)|s_{t+1}).
\label{eq:stoch_db2}
\end{equation}
Note that an odd state can lead to many possible next even states due to stochasticity in the environment.
With the introduction of odd states, we isolate the effect of choosing the action to apply in the environment and of the stochastic state transition given an action, with a deterministic and a stochastic step.

\paragraph{Training a Stochastic GFlowNet.} Combining the two steps, we obtain a novel flow consistency constraint which we call \emph{stochastic GFlowNets} based on detailed balance (DB), where $P$ denotes the state transition function: 
\begin{equation}
\begin{split}
 &F(s_t) \pi(a_t | s_t) P(s_{t+1}|(s_t, a_t)) \\
= & F(s_{t+1}) \pi_B((s_t, a_t)|s_{t+1}).
\label{eq:stoch_db}
\end{split}
\end{equation}

In practice, for training stochastic GFlowNets, we would minimize the loss $\mathcal{L}_{\rm StochGFN-DB}(s, a, s')$ in Eq.~\eqref{eq:stoch_db_loss} based on the flow consistency constraint from Eq.~\eqref{eq:stoch_db}, which is trained on a log-scale.
\begin{equation}
\begin{split}
& \left( \log F(s_t) + \log \pi(a_t|s_t) + \log {P}(s_{t+1}|(s_t, a_t)) \right.\\
& \left.  - \log F(s_{t+1}) - \log \pi_B((s_t, a_t) | s_{t+1}) \right)^2. 
\end{split}
\label{eq:stoch_db_loss}
\end{equation}
Note that our proposed methodology is general and can be applied to other GFlowNet learning objectives such as trajectory balance (TB), as we discuss in Section~\ref{sec:analysis_discussion}.

\paragraph{Learning the dynamics model.} Since the transition dynamics $P(\cdot|s,a)$ are unknown in general, we need to learn it.
In practice, we learn a model $\hat{P}$ with parameters $\phi$ to approximate $P$ through maximum likelihood estimation (other techniques from generative and dynamics modeling~\citep{venkatraman2015improving} could also be applied).
We optimize its parameters with the interaction data using the Adam optimizer~\citep{kingma2014adam} based on the loss function in Eq.~\eqref{eq:loss_model}, where the output of $\hat{P}$ is a softmax distribution across all possible next states.
The data is sampled from a experience replay buffer, which stores interaction data $\{s, a, s'\}$ from the GFlowNet policy and the environment.
\begin{equation}
\mathcal{L}_{\rm model}(s,a,s') = -\log \hat{P}(s'|s,a)
\label{eq:loss_model}
\end{equation}

\begin{figure}[!h]
\centering
\includegraphics[width=0.6\linewidth]{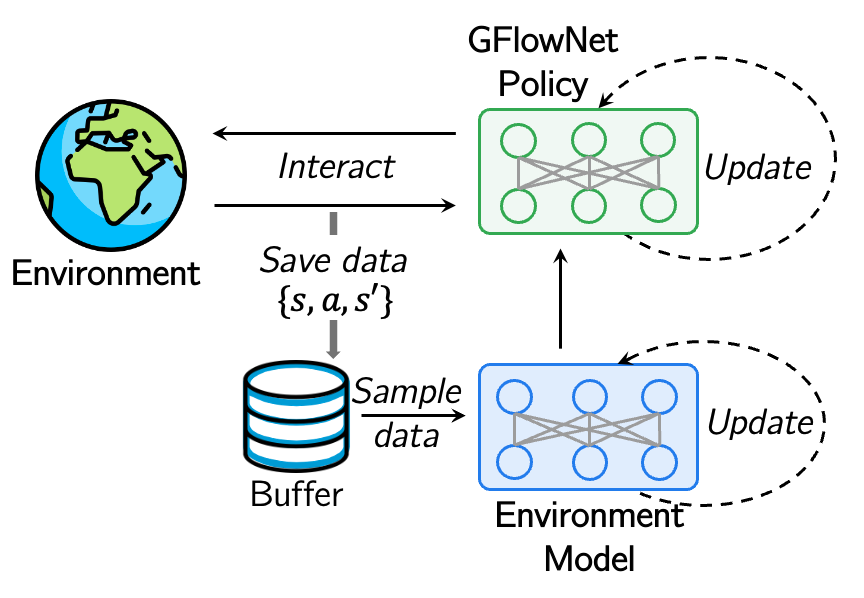}
\caption{Illustration of Stochastic GFlowNets.}
\label{fig:illustration}
\end{figure}

\paragraph{Practical implementation.} Figure~\ref{fig:illustration} illustrates the major components of Stochastic GFlowNets as described above and how they interact with each other. The procedure for training Stochastic GFlowNet based on DB is summarized in Algorithm~\ref{alg}.

\begin{algorithm}[!h]
\caption{Stochastic Generative Flow Networks}
\begin{algorithmic}[1]
\STATE Initialize the forward and backward policies $\pi$, $\pi_{B}$, and the state flow function $F$ with parameters $\theta$\\ 
\STATE Initialize the transition dynamics $\hat{P}$ with parameters $\phi$
\STATE Initialize experience replay buffer $\mathcal{B}$ \\
\FOR {each training step $t=1$ to $T$}
\STATE Collect a batch of $M$ trajectories $\tau=\{s_0 \to \cdots \to s_n\}$ from the policy $\pi$, and store them in $\mathcal{B}$ \\
\STATE Update the stochastic GFN model according to the loss $\mathcal{L}_{\rm StochGFN-DB}$ in Eq.~\eqref{eq:stoch_db_loss} based on $\{\tau\}_{i=1}^M$\\
\STATE Sample a batch of $K$ trajectories from $\mathcal{B}$ \\
\STATE Update the transition dynamics model according to the loss $\mathcal{L}_{\rm model}$ in Eq.~\eqref{eq:loss_model} using data sampled from the replay buffer \\
\ENDFOR
\end{algorithmic}
\label{alg}
\end{algorithm}

\subsection{Discussion on the Applicability to Trajectory Balance (TB)}
\label{sec:analysis_discussion}
As discussed in Section~\ref{sec:method}, our proposed method is versatile and can be applied to other GFlowNet learning objectives beyond DB.
We state the flow consistency constraint for Stochastic TB in Eq.~\eqref{eq:stoch_tb}, which is obtained via a telescoping calculation based on Eq.~\eqref{eq:stoch_db}.
\begin{equation}
Z \prod_{t=0}^{n-1} \pi(a_t | s_t) P(s_{t+1}|(s_t, a_t))
= R(x) \prod_{t=0}^{n-1} \pi_B((s_t, a_t)|s_{t+1})
\label{eq:stoch_tb}
\end{equation}
In practice, we can train with Stochastic TB by minimizing the loss $\mathcal{L}_{\rm StochGFN-TB}(s, a, s')$ obtained from Eq~\eqref{eq:stoch_tb}, \ie,
\begin{equation}
\begin{split}
& \bigg[ \log Z + \sum_{t=0}^{n-1} \log \pi(a_t|s_t) +  \sum_{t=0}^{n-1} \log \hat{P}(s_{t+1}|(s_t, a_t))\\
&\quad\quad\quad\quad - \log R(x) -  \sum_{t=0}^{n-1} \log \pi_B((s_t, a_t) | s_{t+1}) \bigg]^2. 
\end{split}
\label{eq:stoch_tb_loss}
\end{equation}
However, as TB is optimized based on a sampled trajectory instead of each transition, it can lead to a larger variance as studied in~\citet{madan2022learning} even in deterministic environments. This problem can be further exacerbated in stochastic environments. In Section~\ref{sec:exp_stoch_tb}, we find that the Stochastic TB underperforms relative to Stochastic DB, presumable due to a larger variance, as studied by \citet{madan2022learning}. 

\section{Experiments}
In this section, we conduct extensive experiments to investigate the following key questions: i) How much can Stochastic GFNs improve over GFNs in the presence of stochastic transition dynamics?
ii) Can Stochastic GFNs be built upon different GFlowNets learning objectives?
iii) Can Stochastic GFNs scale to the more complex and challenging tasks of generating biological sequences?

\subsection{GridWorld}
\subsubsection{Experimental Setup} \label{sec:grid_setup}
\begin{figure}[!h]
\centering
\includegraphics[width=0.35\linewidth]{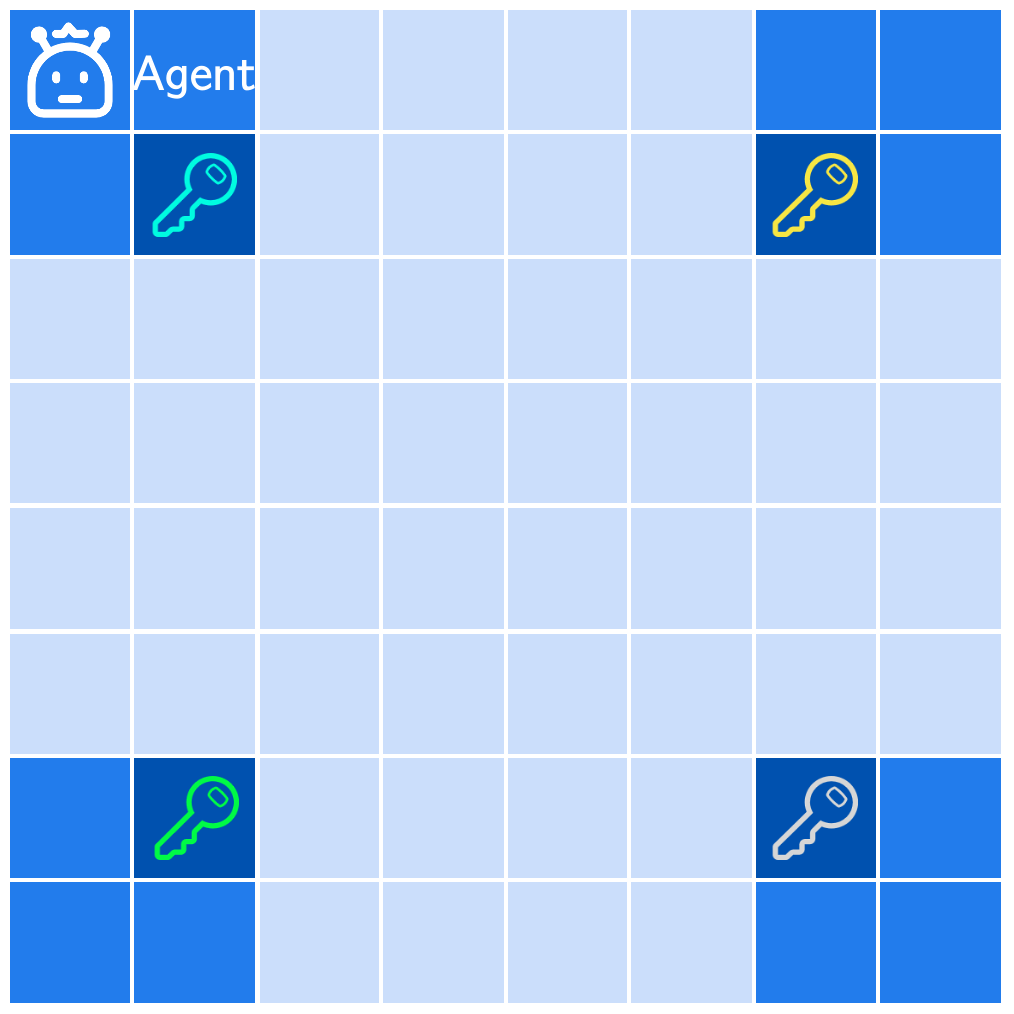}
\caption{The GridWorld environment. The agent starts at the top-left corner and reward is largest at the four dark blue positions near the four corners (with the keys), lower in the $2 \times 2$ squares near the corner, and yet lower in other (light blue) positions. This can be extended to different sizes $H$, as well as different degrees of noise $\alpha$ in the (state, action)-to-state transitions.}
\label{fig:grid_env}
\end{figure}

We first conduct a series of experiments in the GridWorld task introduced in~\citet{bengio2021flow} to understand the effectiveness of Stochastic GFlowNets. 
An illustration of the task with size $H \times H$ is shown in Figure~\ref{fig:grid_env}.
At each time step, the agent takes an action to navigate in the grid, where possible actions include operations to increase one coordinate and also a stop operation to terminate the episode, ensuring the underlying Markov decision process (MDP) is a directed acyclic graph.
The agent obtains a reward $R(x)$ as defined in~\citet{bengio2021flow} when the trajectory ends at a terminal state $x$. The reward function $R(x)$ has $4$ modes located at the corners of the map as illustrated in Figure~\ref{fig:grid_env}.
The goal for the agent is to model the target reward distribution, and captures all the modes of the reward function.
The shade of color in Figure~\ref{fig:grid_env} indicates the magnitude of rewards, where a darker color corresponds to a larger reward.
We consider a variant with stochastic transition dynamics, where the randomness in the environment is injected following~\citet{machado2018revisiting,yang2022dichotomy} in GridWorld and all other benchmark tasks in Sections~\ref{sec:bit}-\ref{sec:amp}. Specifically, the environment transitions according to the selected action with probability $1-\alpha$, while with probability $\alpha$ the environment executes 
a uniformly chosen action (like slipping to its neighboring regions randomly in Figure~\ref{fig:grid_env}).

\begin{figure*}[!h]
\centering
\subfloat[Small.]{\includegraphics[width=0.28\linewidth]{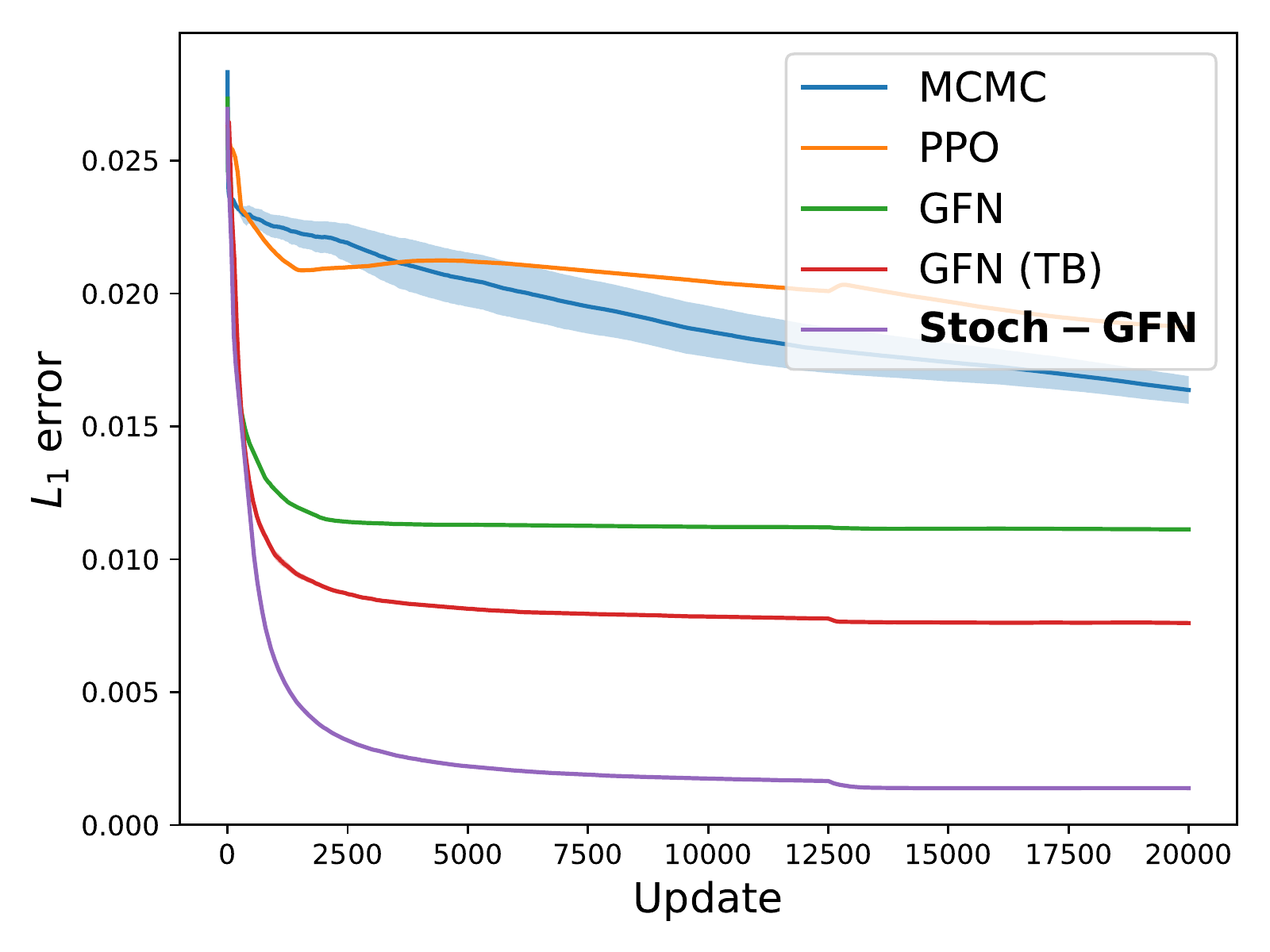}}
\subfloat[Medium.]{\includegraphics[width=0.28\linewidth]{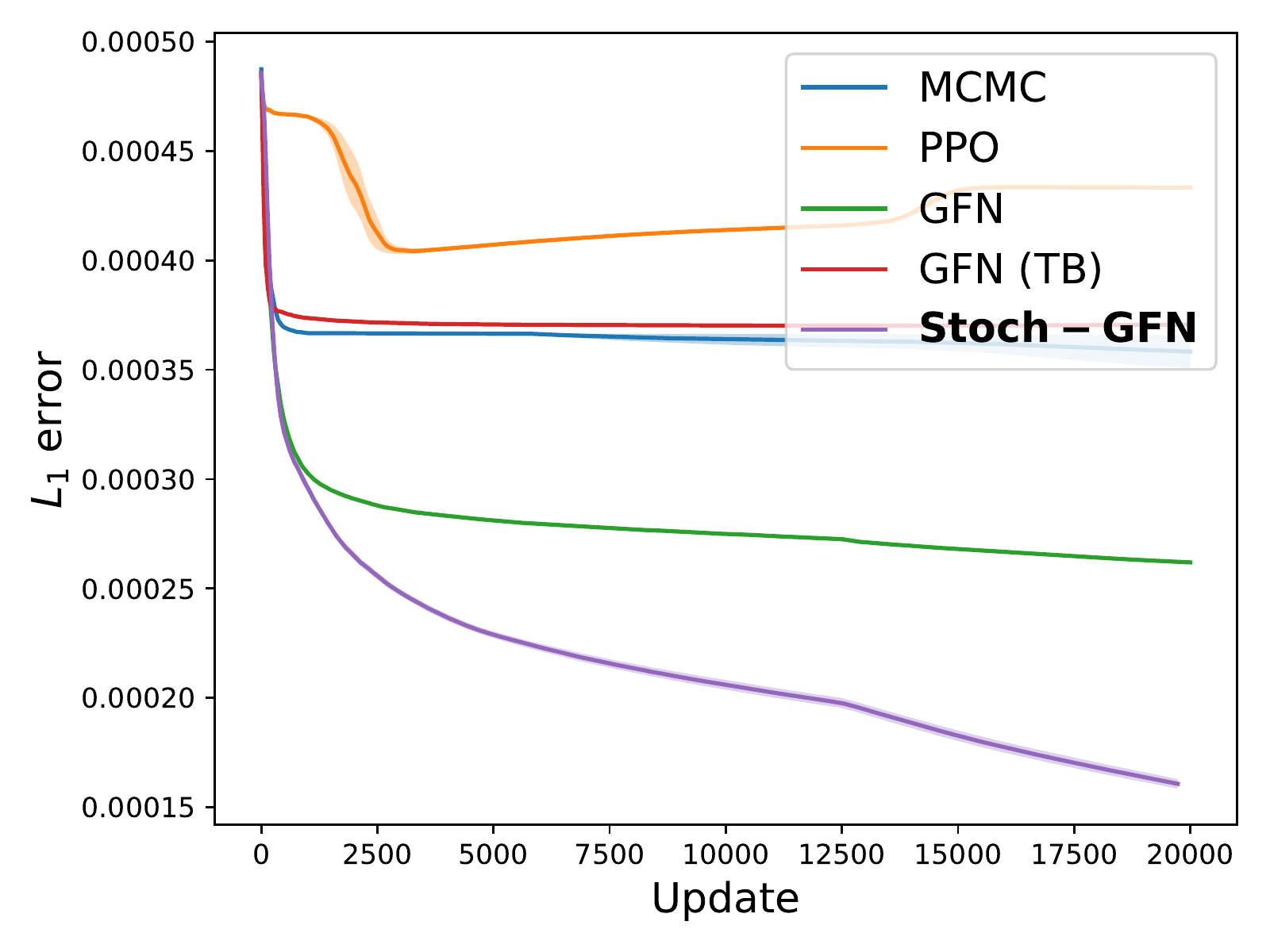}}
\subfloat[Large.]{\includegraphics[width=0.28\linewidth]{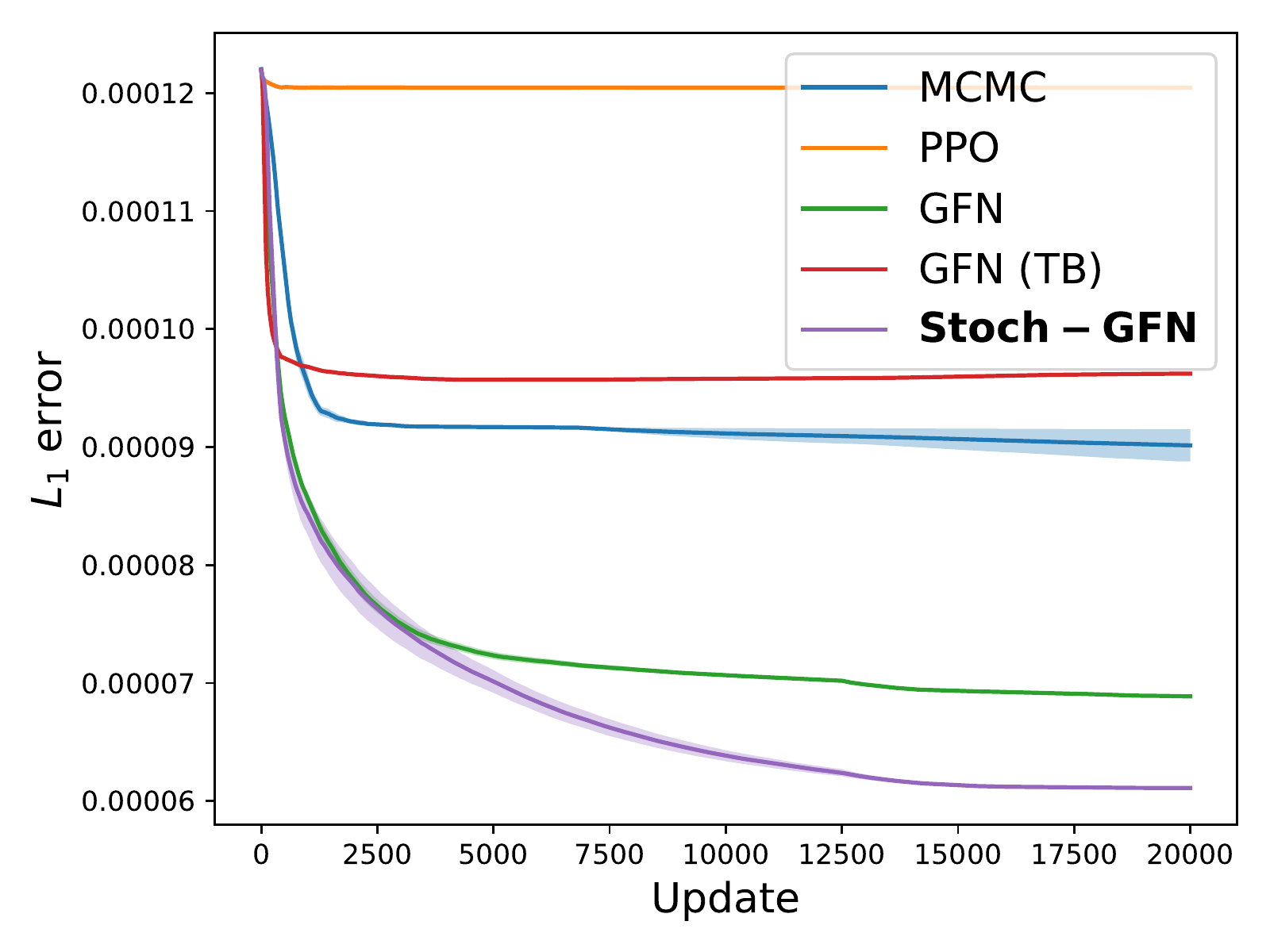}}
\caption{Comparison results of $L_1$ error in GridWorld with increasing sizes of the map.}
\label{fig:grid_vary_size}
\end{figure*}

\begin{figure*}[!h]
\centering
\subfloat[Small.]{\includegraphics[width=0.28\linewidth]{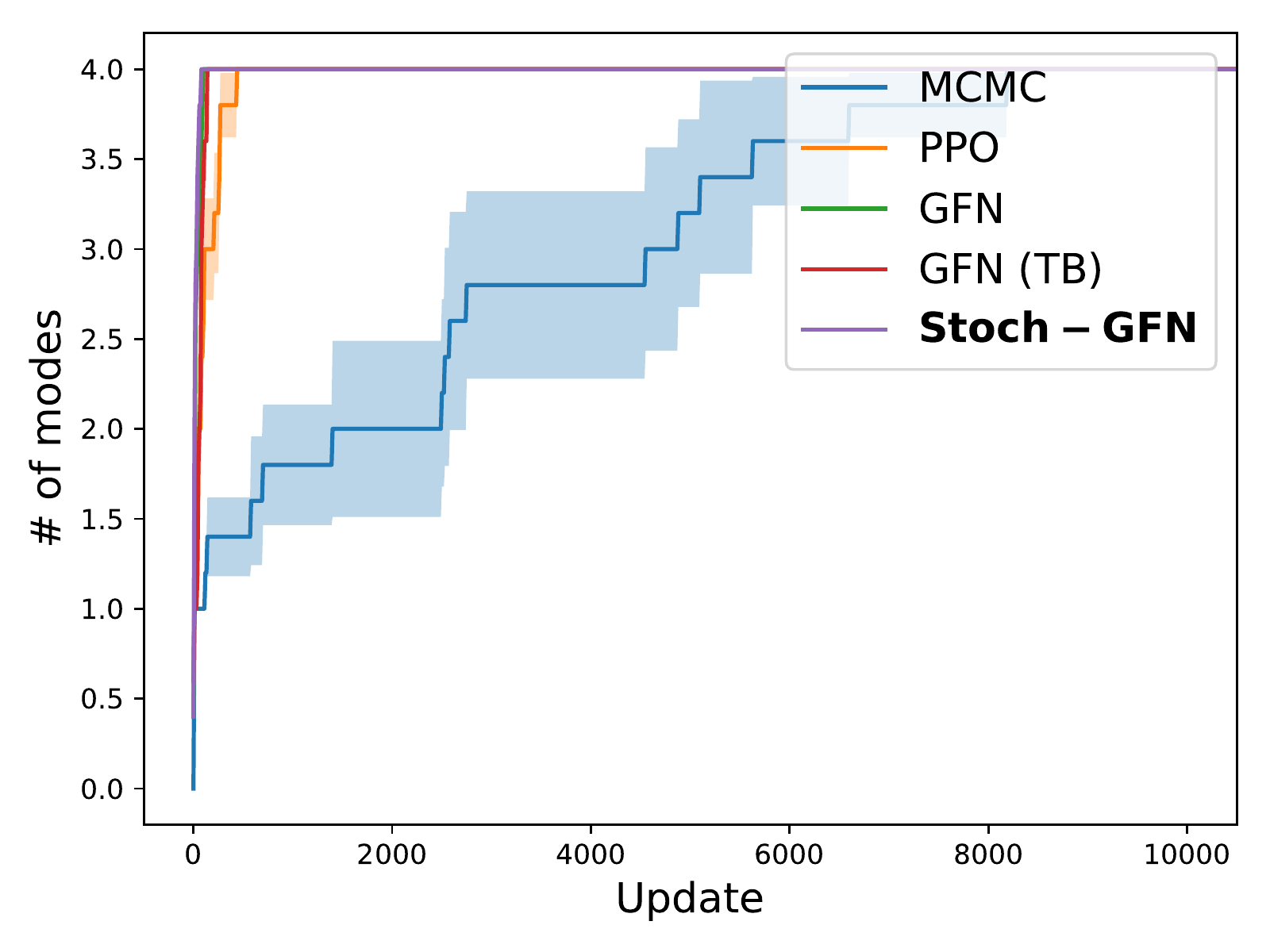}}
\subfloat[Medium.]{\includegraphics[width=0.28\linewidth]{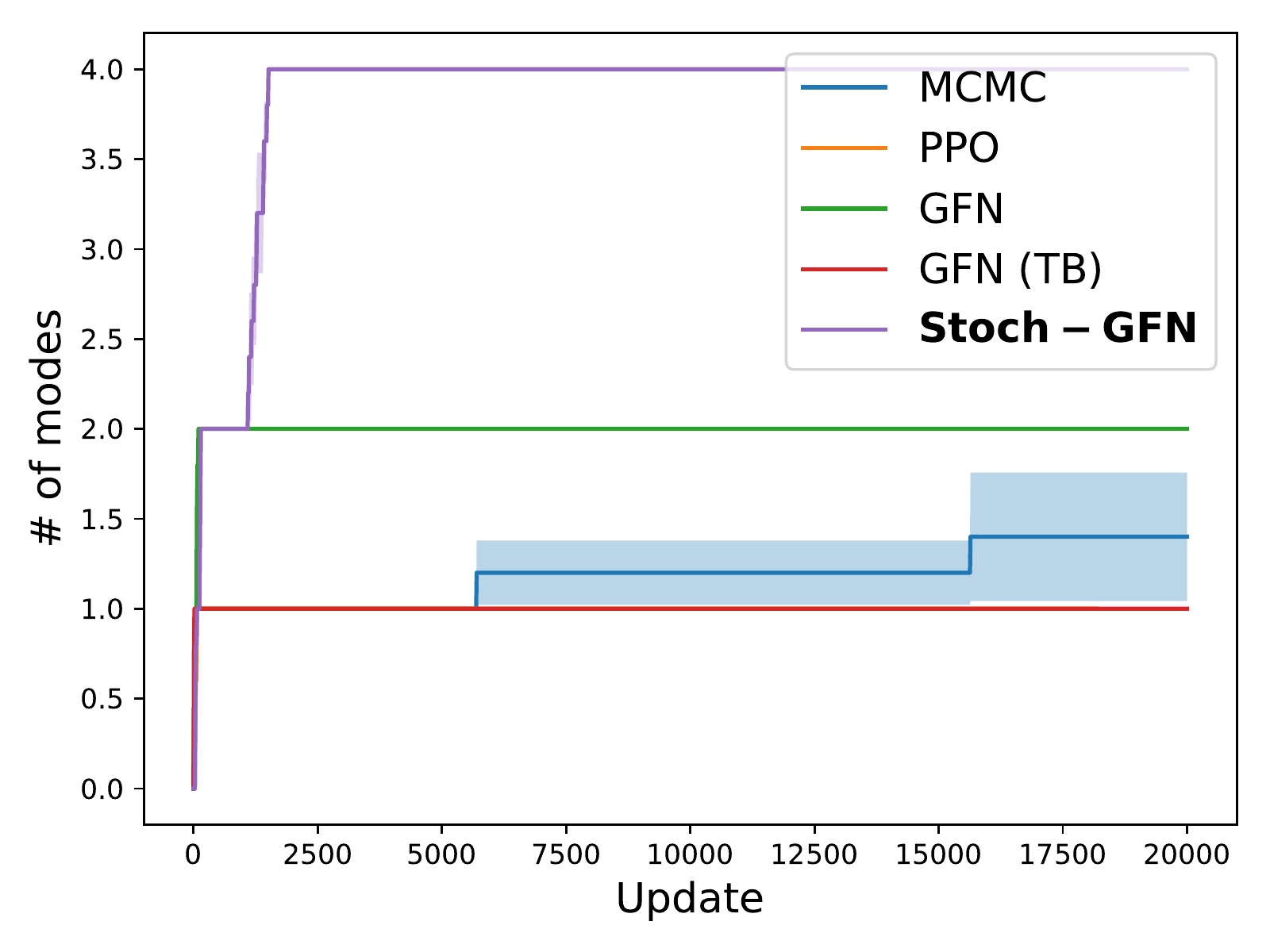}}
\subfloat[Large.]{\includegraphics[width=0.28\linewidth]{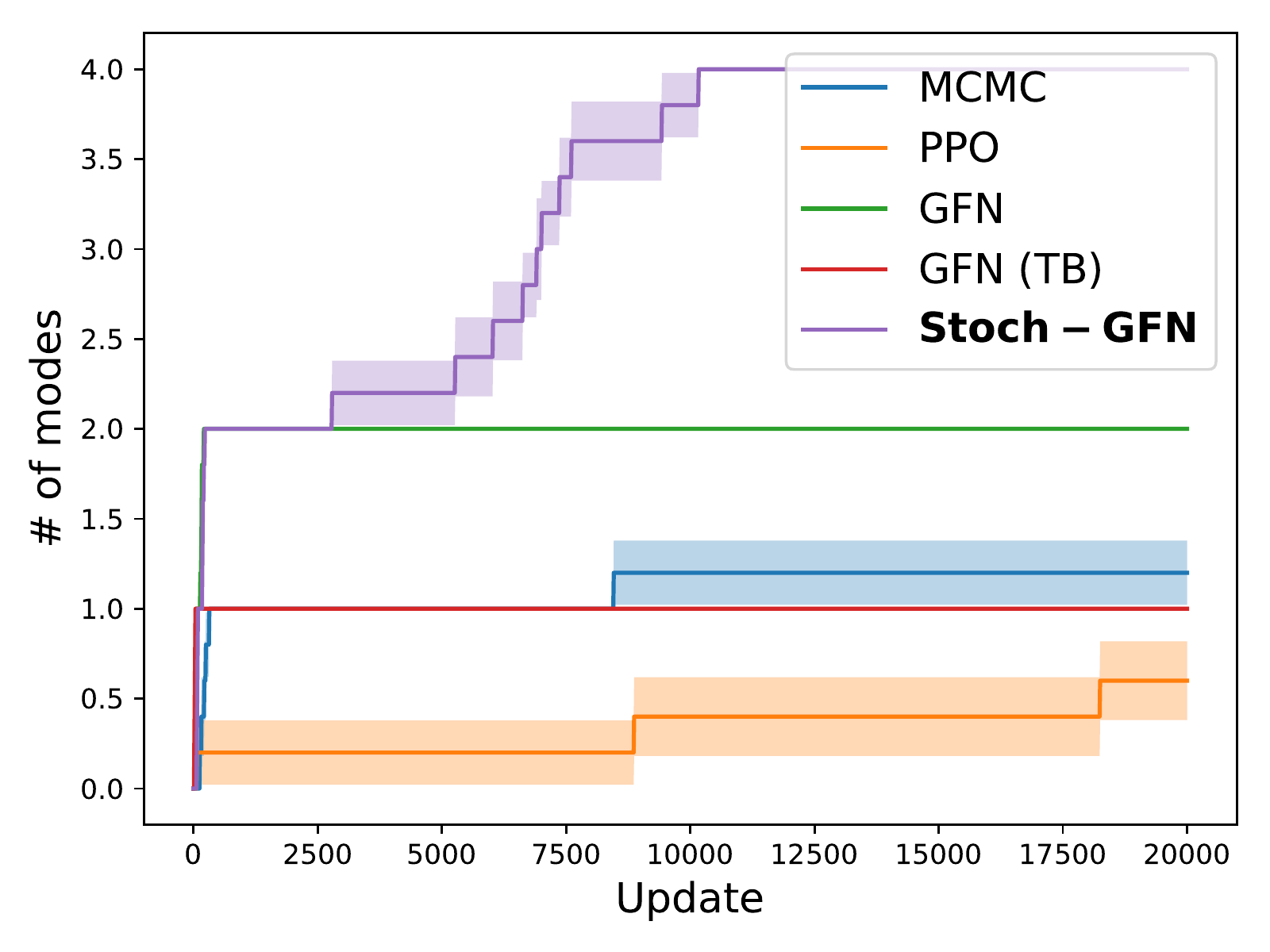}}
\caption{Comparison results of the number of modes captured during the training process in GridWorld with increasing sizes of the map.}
\label{fig:grid_vary_size_modes}
\end{figure*}

We compare Stochastic GFlowNet against vanilla GFlowNets trained with detailed balance (DB)~\citep{bengio2021foundations} and trajectory balance (TB)~\citep{malkin2022trajectory} learning objectives, Metropolis-Hastings-MCMC~\citep{xie2021mars}, and PPO~\citep{schulman2017proximal} methods. 
We evaluate each method in terms of the empirical $L_1$ error defined as $\mathbb{E}[|p(x)-\pi(x)|]$, with $p(x)=\frac{R(x)}{Z}$ denoting the true reward distribution, and we estimate $\pi$ according to repeated sampling and calculating frequencies for visiting every possible state $x$.
We also compare them in terms of the number of modes discovered by each method during the course of training.
Each algorithm is run for $5$ different seeds, and the performance is reported in its mean and standard deviation.
We implement all baselines based on the open-source code\footnote{\url{https://github.com/GFNOrg/gflownet}}, and a detailed description of the hyperparameters and setup can be found in Appendix A.1.

\subsubsection{Performance Comparison}
We now study the effectiveness of Stochastic GFNs on small, medium, and large GridWorlds with increasing sizes $H$, and different levels of stochasticity.

\paragraph{Varying sizes of the map.} Figure~\ref{fig:grid_vary_size} demonstrates the empirical $L_1$ error for each method in GridWorld (with a stochasticity level of $\alpha=0.25$) with increasing sizes.
As shown, MCMC does not perform well and PPO fails to converge.
We also observe that the performance of TB gets much worse as the size of the problem increases, which may be attributed to a larger gradient variance~\citep{madan2022learning}.
Stochastic GFlowNets significantly outperform the baselines, and converge fastest and to the lowest empirical $L_1$ error.
Figure~\ref{fig:grid_vary_size_modes} illustrates the number of modes discovered by each method during the course of training. As demonstrated, in stochastic environments (where the original convergence guarantees of GFlowNets do not hold), existing GFlowNet methods including DB and TB fail to discover all of the modes in maps with larger sizes.
It is also worth noting that TB performs much worse than DB in terms of the number of modes discovered with increasing sizes of the maps, as it is optimized on the trajectory level with a sampled trajectory instead of the transition level as in DB, and can induce large variance.
The proposed Stochastic GFlowNet method outperforms previous GFlowNet methods as well as MCMC and PPO by a large margin, while being able to efficiently discover different modes in maps with different sizes.

\begin{figure}[!h]
\centering
\subfloat[$\alpha=0.5$.]{\includegraphics[width=0.5\linewidth]{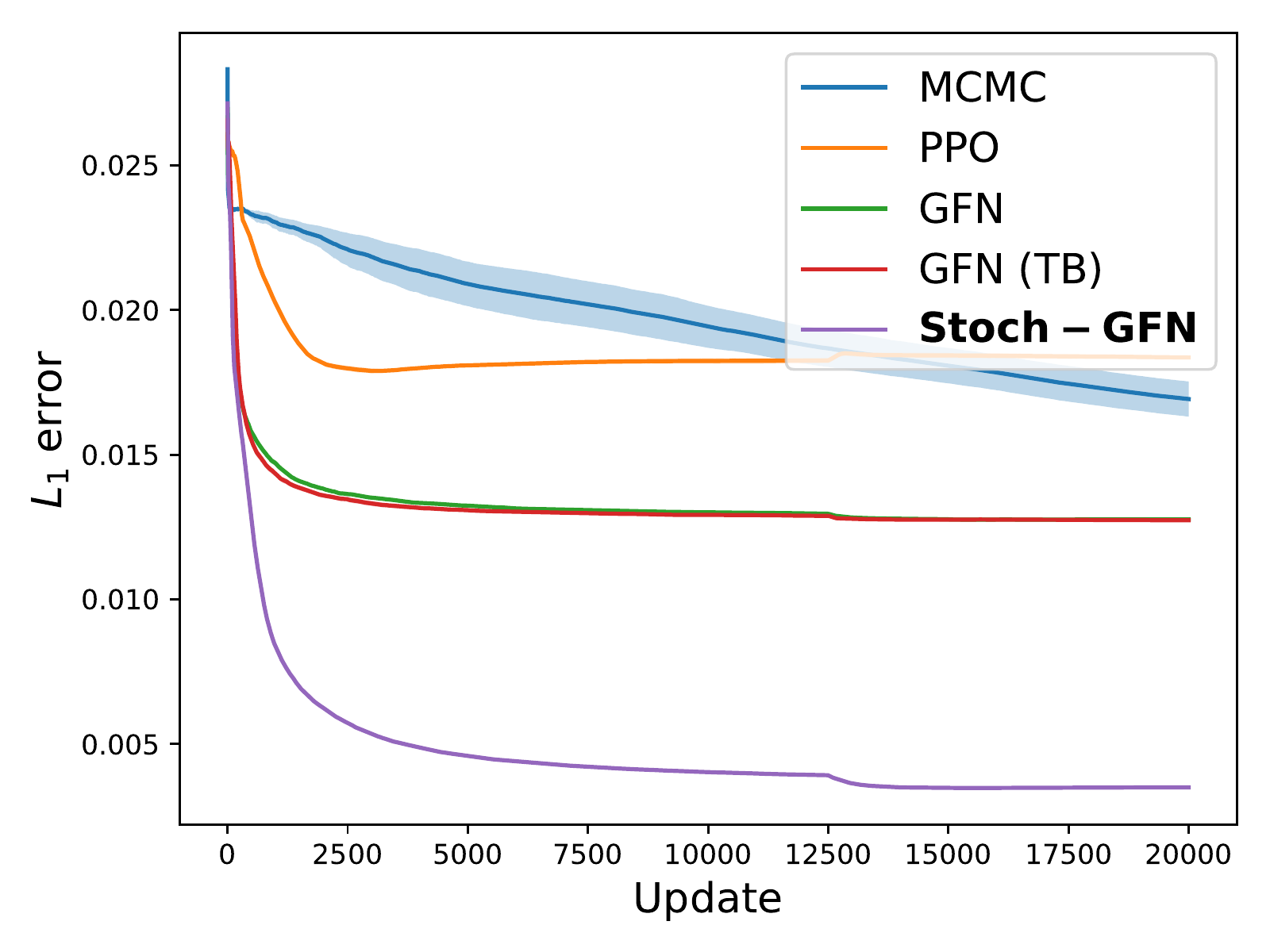}}
\subfloat[$\alpha=0.9$.]{\includegraphics[width=0.5\linewidth]{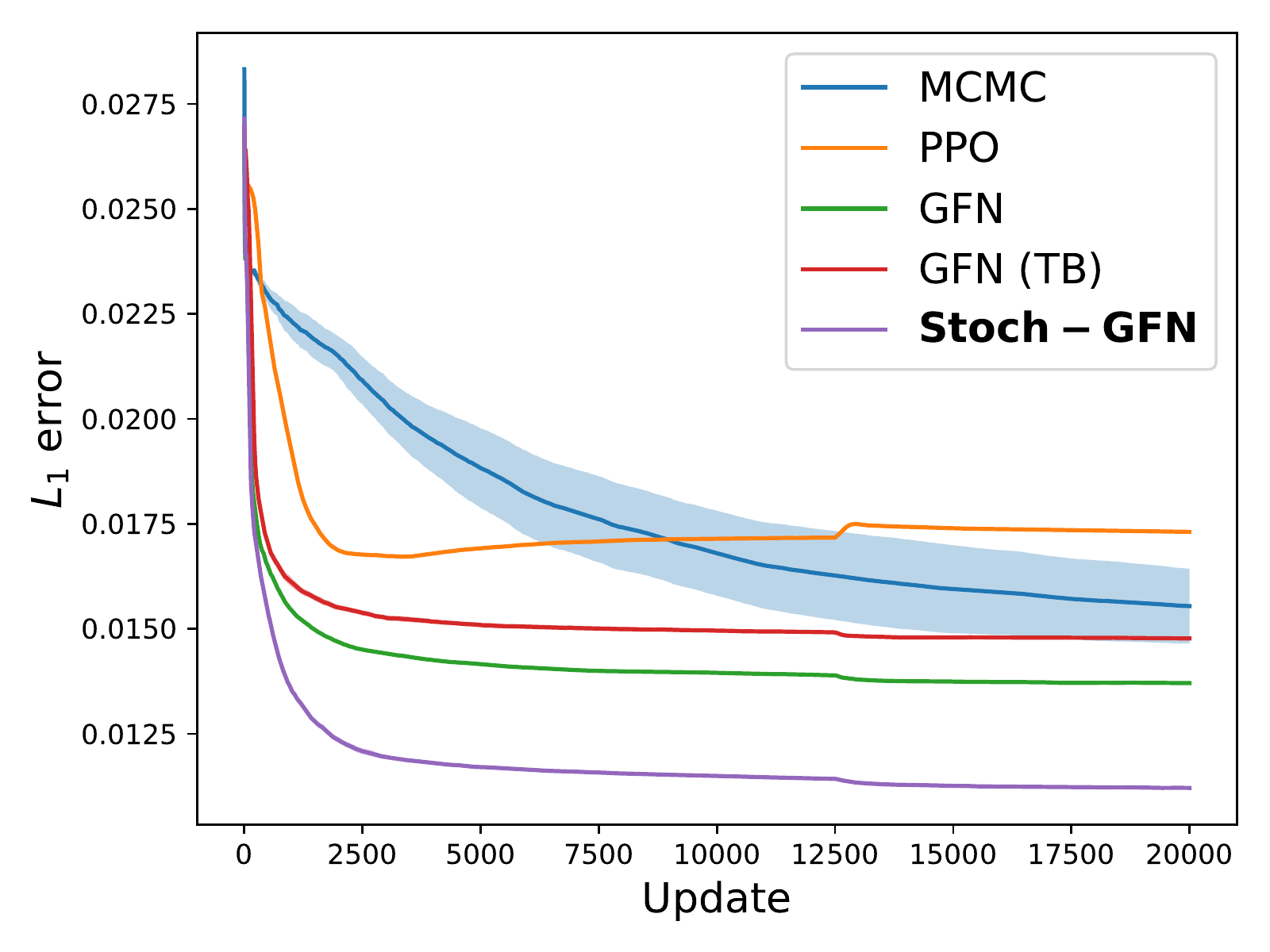}}
\caption{Results in small GridWorld with increasing stochasticity levels $\alpha$.}
\label{fig:grid_vary_alpha}
\end{figure}

\paragraph{Varying stochasticity levels.} In Figure~\ref{fig:grid_vary_alpha}, we compare different methods in a small GridWorld with an increasing level of stochasticity $\alpha$.
We observe that TB also fails to learn well with an increasing $\alpha$, and performs worse than DB besides the decreased performance with increasing sizes.
On the other hand, Stochastic GFlowNets outperform the baselines by a significant margin, and are robust to higher levels of stochasticity, successfully handling stochastic transition dynamics.

\subsubsection{Compatibility with Different GFlowNet Learning Objectives} \label{sec:exp_stoch_tb}
\begin{figure}[!h]
\centering
\subfloat[]{\includegraphics[width=0.33\linewidth]{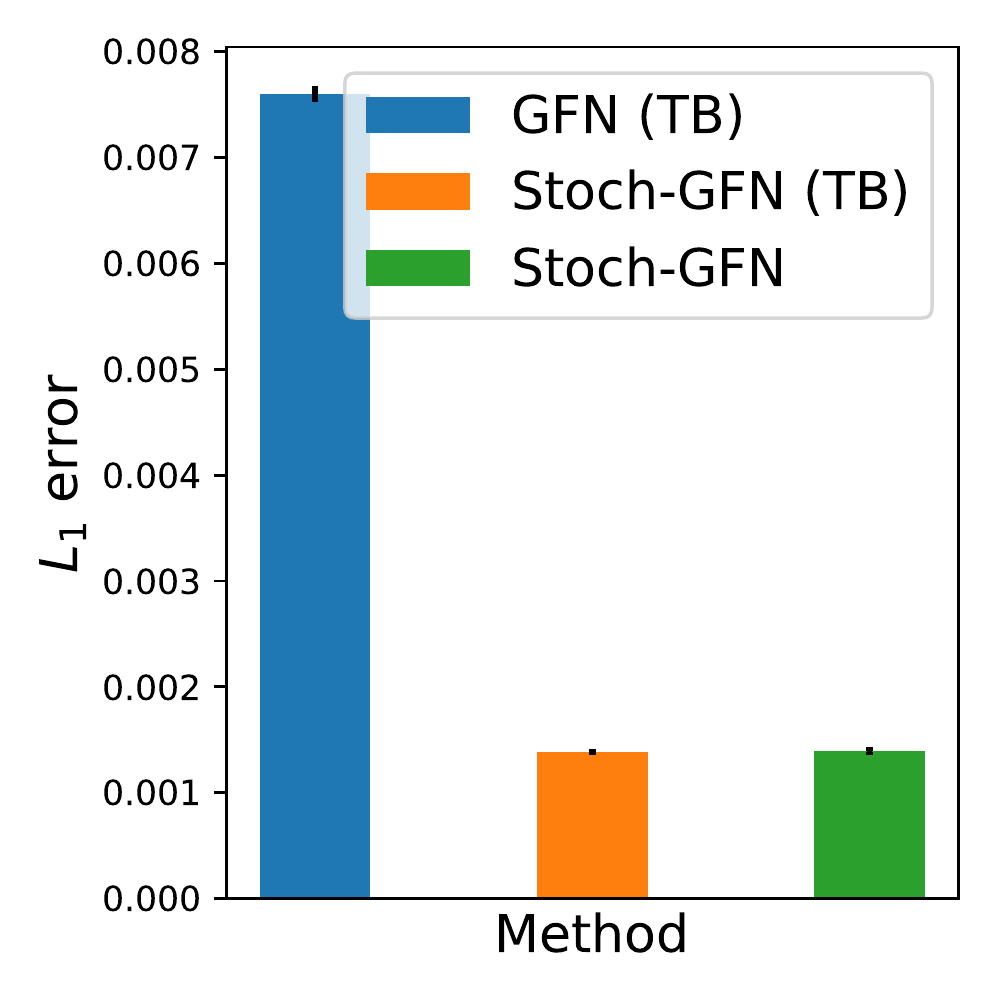}}
\subfloat[]{\includegraphics[width=0.33\linewidth]{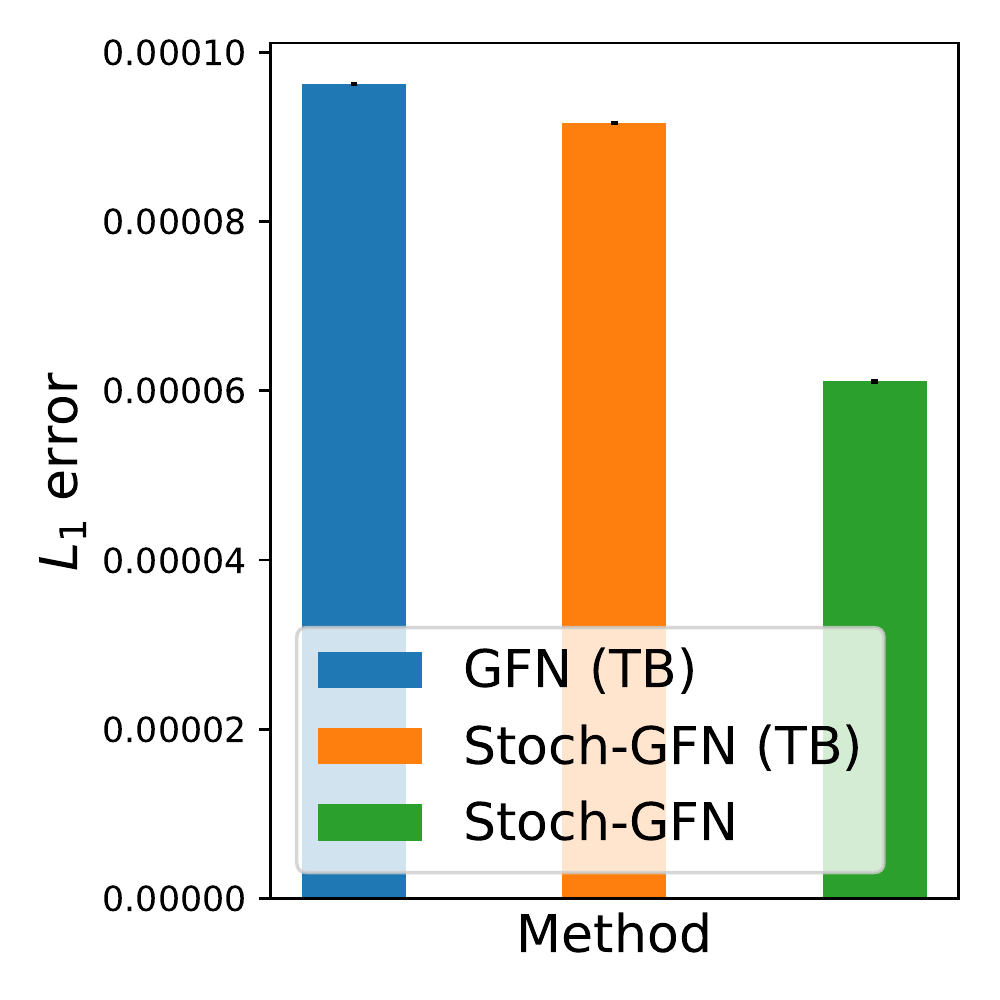}}
\subfloat[]{\includegraphics[width=0.33\linewidth]{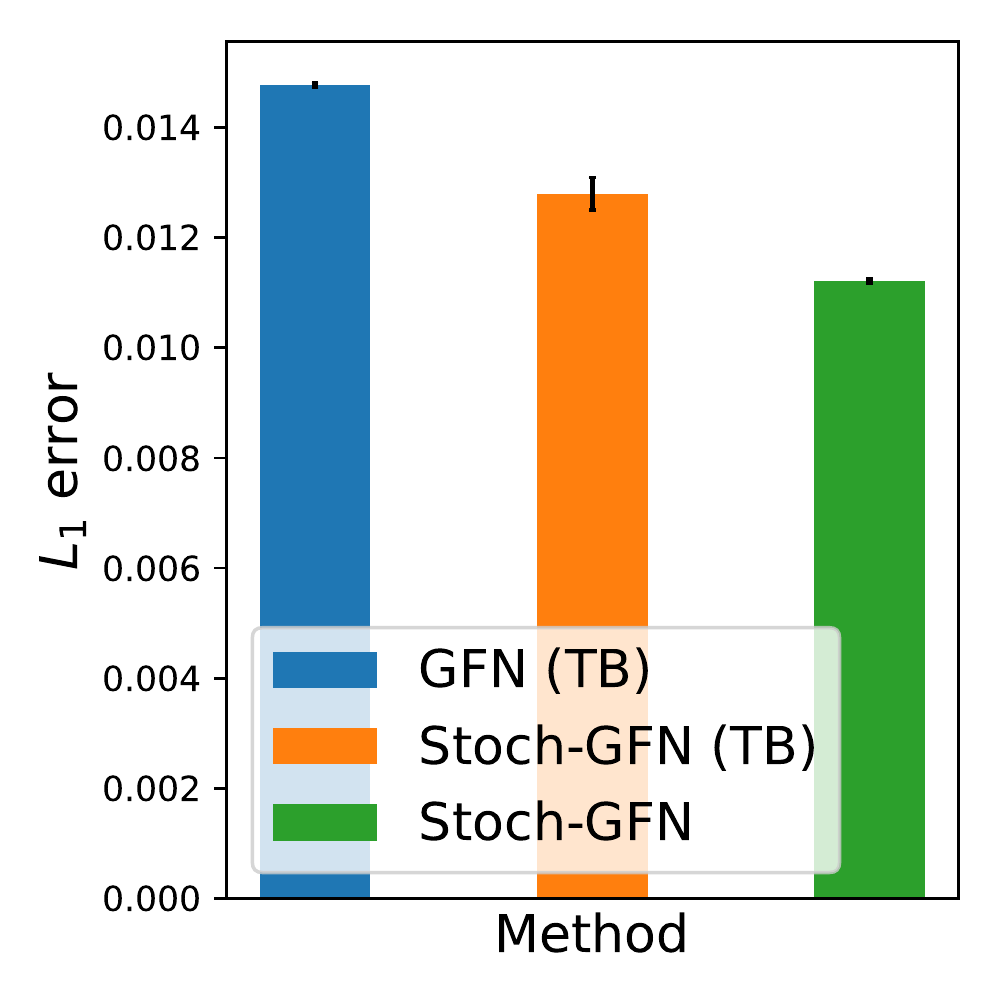}}
\caption{Results of Stochastic GFlowNet when built upon the trajectory balance (TB) objective in GridWorld with increasing sizes $H$ and stochasticity levels $\alpha$. (a) Small, low stochasticity level. (b) Large, low stochasticity level. (c) Small, high stochasticity level.}
\label{fig:grid_tb}
\end{figure}

In this section, we study Stochastic GFlowNets with the trajectory balance (TB) objective as described in Section~\ref{sec:analysis_discussion}. 
We evaluate Stochastic TB in GridWorlds with different sizes (including small with $H=8$ and large with $H=128$) and stochasticity levels (including low with $\alpha=0.25$ and high with $\alpha=0.9$). Specifically, Figure~\ref{fig:grid_tb}(a) corresponds to the result in a small map with a low stochasticity level, Figure~\ref{fig:grid_tb}(b) illustrates the results in a large map with a low stochasticity level, while Figure~\ref{fig:grid_tb} shows the results in a small map with a high stochasticity level.

\begin{figure*}[!t]
\centering
\subfloat[]{\includegraphics[width=0.28\linewidth]{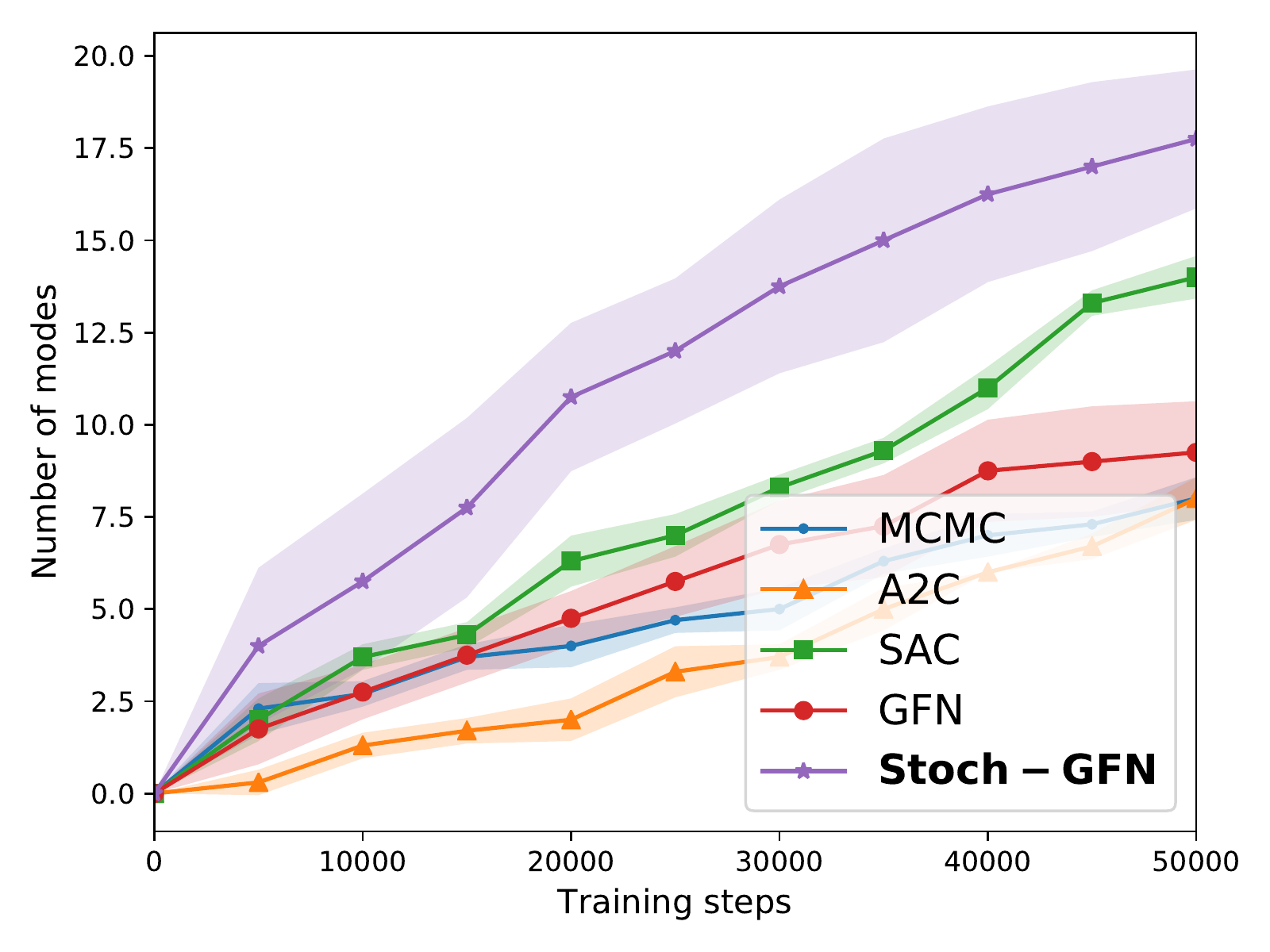}}
\subfloat[]{\includegraphics[width=0.28\linewidth]{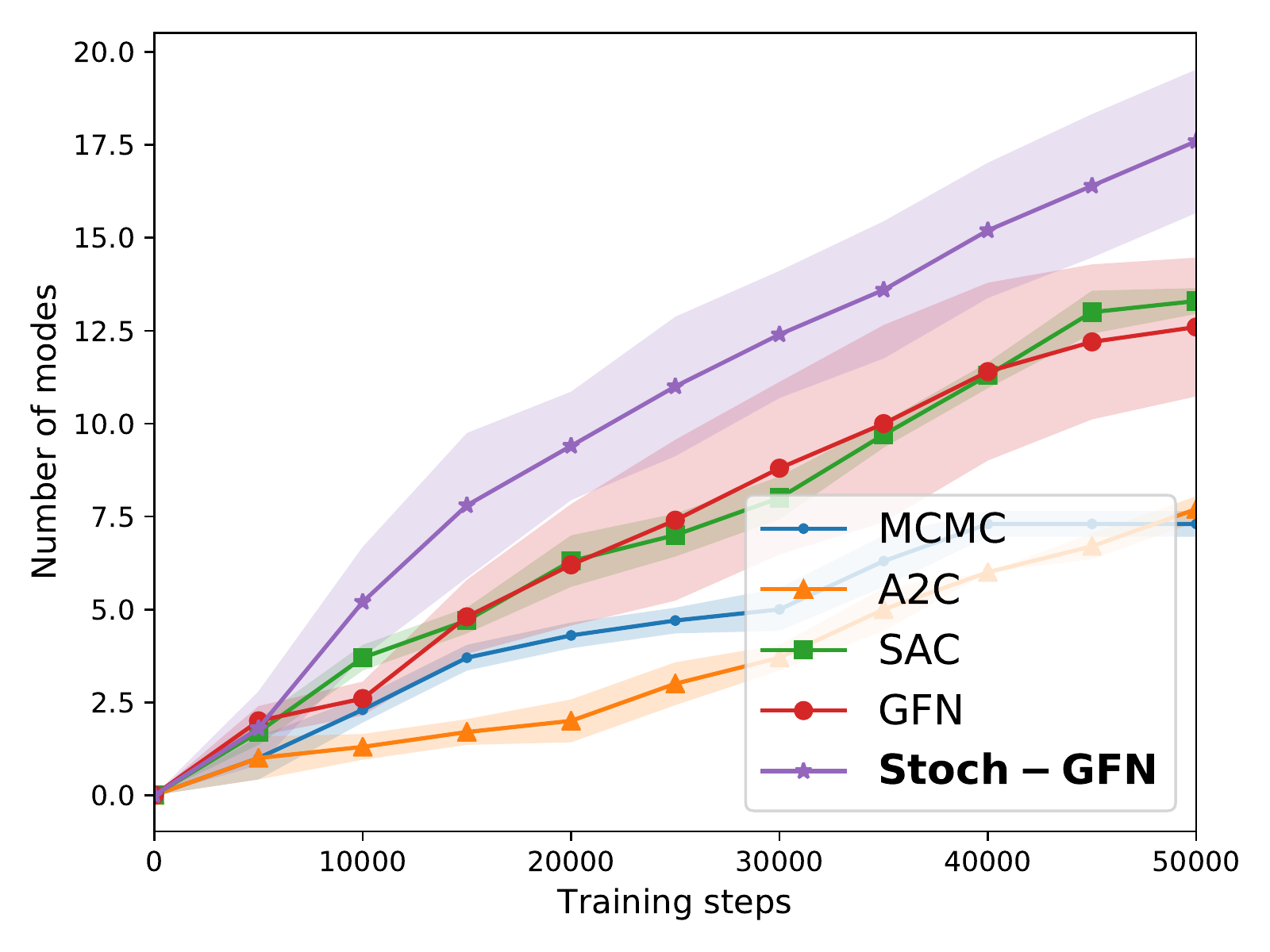}}
\subfloat[]{\includegraphics[width=0.28\linewidth]{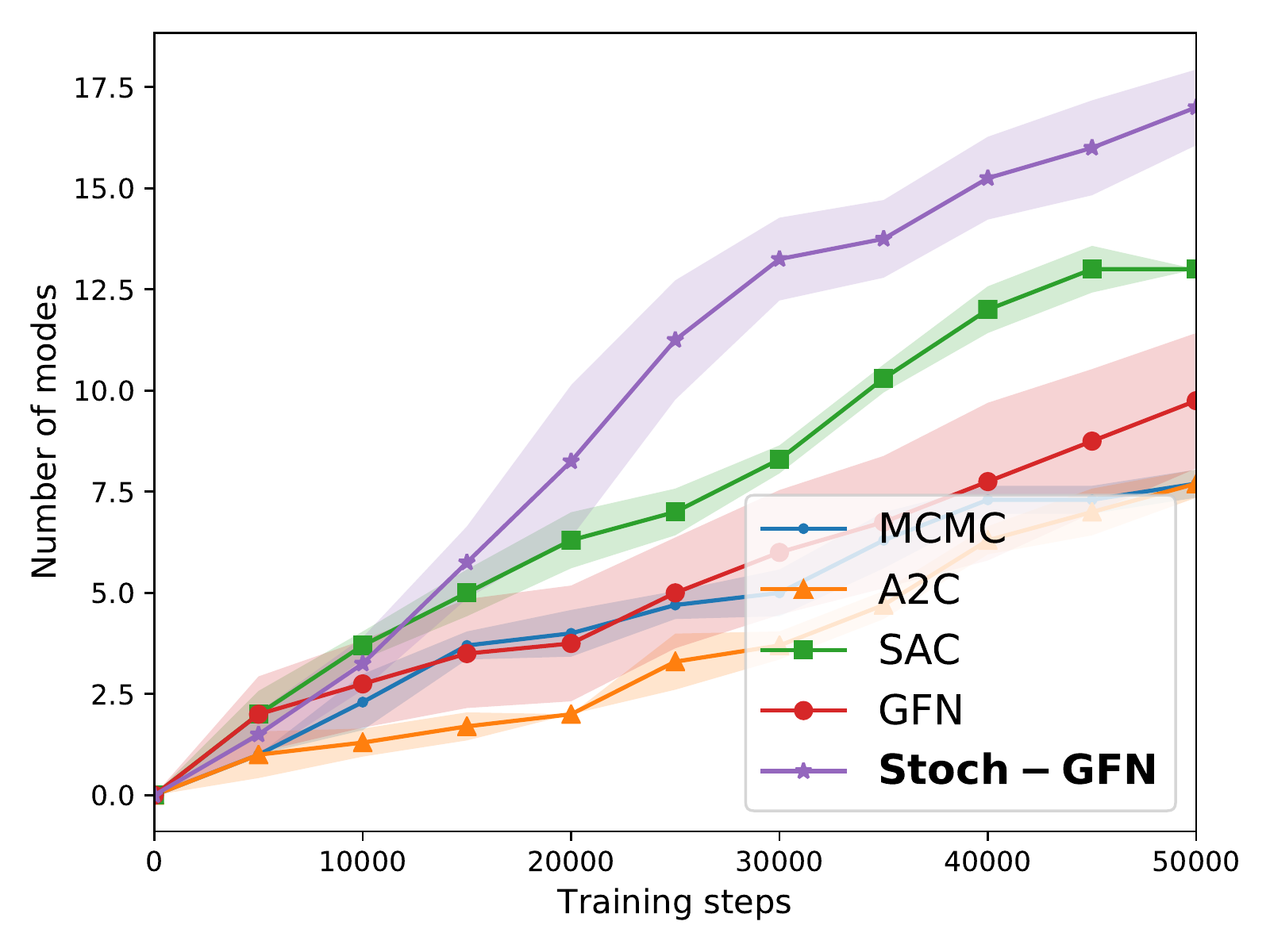}}\\
\subfloat[]{\includegraphics[width=0.28\linewidth]{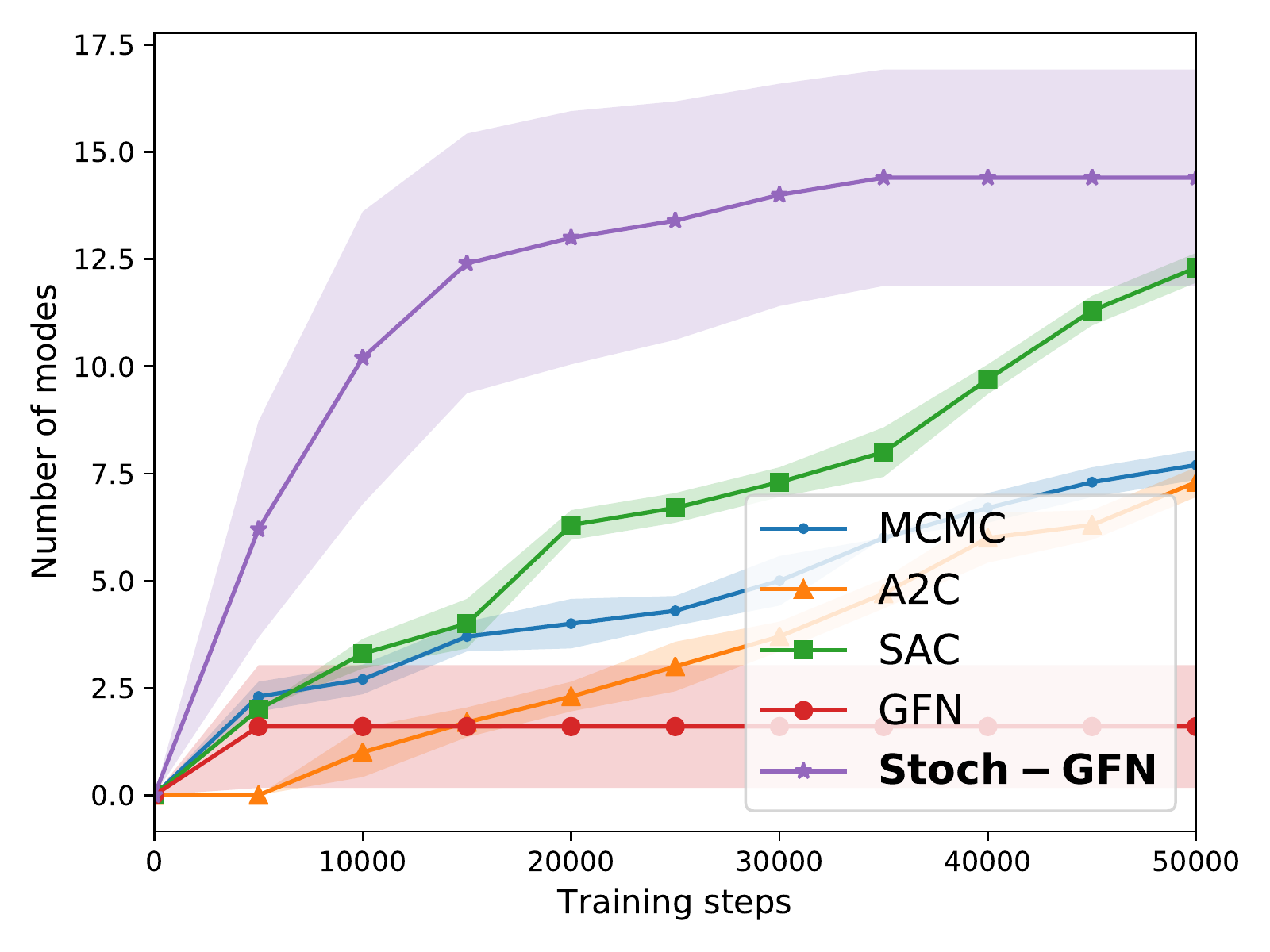}}
\subfloat[]{\includegraphics[width=0.28\linewidth]{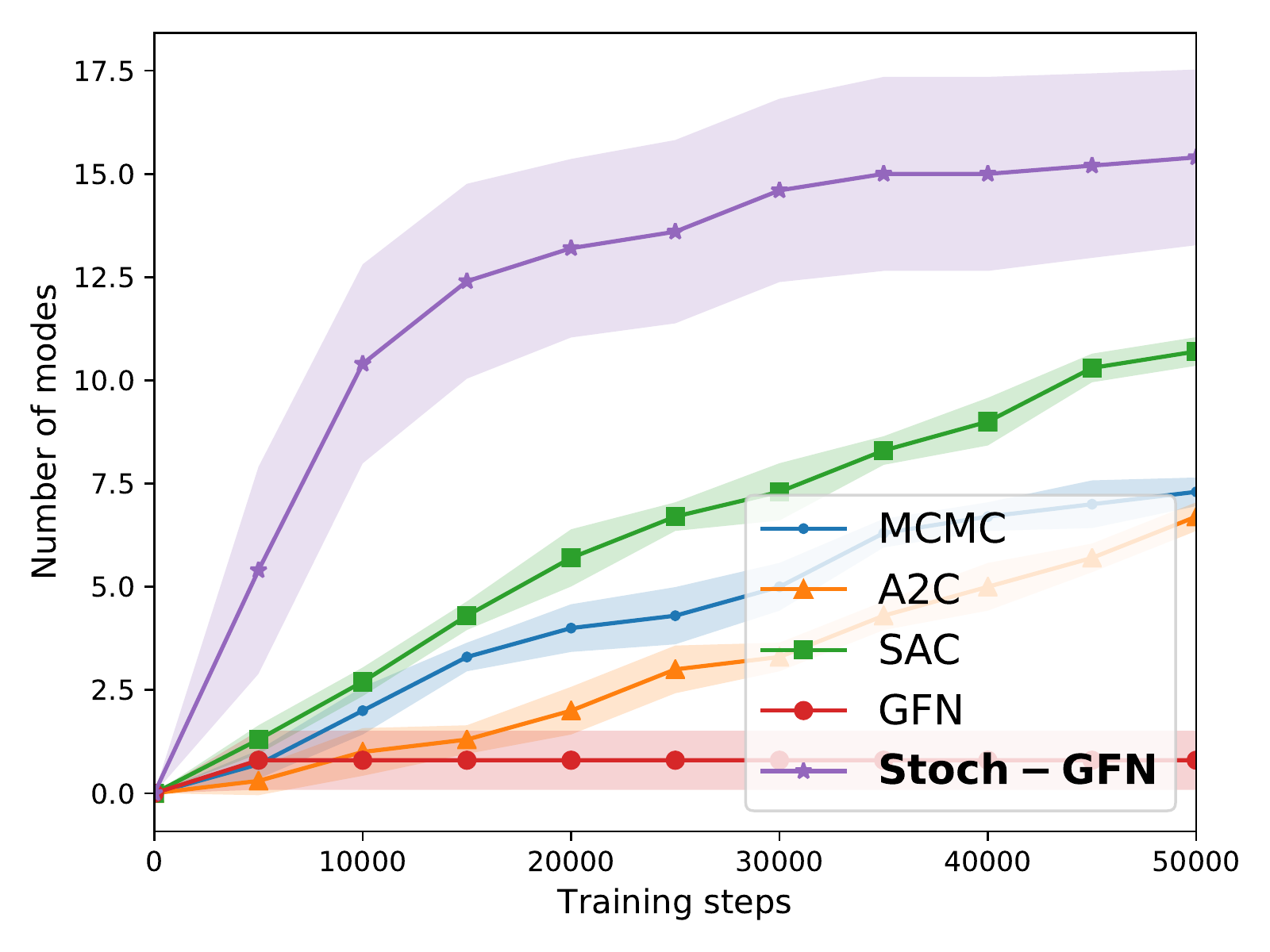}}
\subfloat[]{\includegraphics[width=0.28\linewidth]{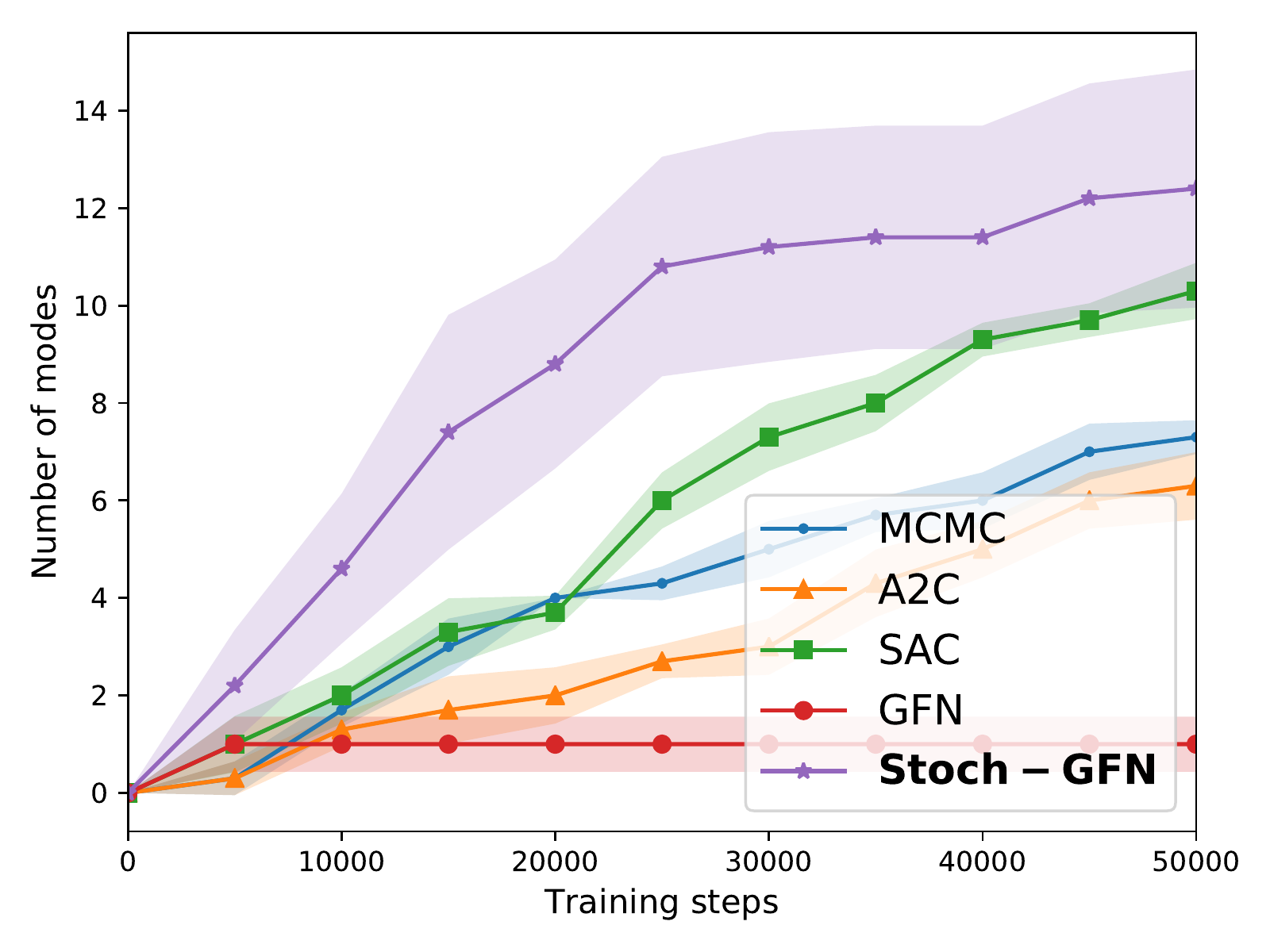}}
\caption{Results in the bit sequence generation task. The first and second rows correspond to the results of the number of bits $k=4$ and $k=2$. The first, second, and third columns correspond to the results of different stochasticity levels of $0.1$, $0.3$, and $0.5$, respectively.}
\label{fig:bit}
\end{figure*}

As shown in Figure~\ref{fig:grid_tb}, Stochastic TB (abbreviated as Stoch-GFN (TB) in the figure) greatly improves the performance of TB, validating the effectiveness of our proposed methodology.
However, we observe that it underperforms relative to Stochastic DB when the scale of the problem increases or with a higher level of stochasticity (Figure~\ref{fig:grid_tb}(c)), which can be attributed to the larger variance of TB~\citep{madan2022learning} in stochastic environments.

\subsection{Autoregressive Sequence Generation} \label{sec:seq}
In this section, we study Stochastic GFN on autoregressive sequence generation tasks~\citep{malkin2022trajectory}.
We first consider a bit sequence generation task to investigate the effect of the size of the action space and length of the trajectory with varying levels of environment stochasticity.
We then study the more realistic and complex tasks of generating biological sequences.

\subsubsection{Bit Sequences} \label{sec:bit}
\paragraph{Task.} In the bit sequence generation task~\citep{malkin2022trajectory}, the agent aims to generate bit sequences of length $n=120$.
At each step, the agent appends a $k$-bit ``word" from a vocabulary $V$ to the current state from left to right, which is a partial sequence.
Note that we consider a stochastic variant of the task, with noise level $\alpha$ as described in Section~\ref{sec:grid_setup}.
The resulting action space has a size of $|V|=2^k$, and the length of the complete trajectories is $\frac{n}{k}$.
Following~\citet{malkin2022trajectory}, we define the reward function $R(x)$ to have modes at a fixed set of bit sequences $M$ with $R(x)=\exp(- \min_{y \in M} d(x, y))$, where $d$ is the edit distance.
We evaluate each method in terms of the number of modes discovered during the course of training.

We study the performance of Stochastic DB with different levels of stochasticity, and compare it against vanilla DB and strong baselines including Advantage Actor-Critic (A2C)~\citep{mnih2016asynchronous}, Soft Actor-Critic (SAC)~\citep{haarnoja2018soft}, and MCMC~\citep{xie2021mars}.
Each method is run for $3$ different seeds and we report the mean and standard deviation.
More details about the experimental setup in the stochastic bit sequence generation task can be found in Appendix A.2. We use the same hyperparameters and architectures as in~\citet{malkin2022trajectory}.

\paragraph{Results.} 
Figure~\ref{fig:bit} demonstrates the number of modes captured by each method throughout the training process with different levels of stochasticity ranging from $0.1$ to $0.5$, where the first and second rows correspond to the results for $k=4$ and $k=2$, respectively.
We observe that regular GFlowNets (GFN in the figure) fail to learn well, particularly when the trajectories are longer (with a smaller value of $k$).
On the other hand, the Stochastic GFlowNet (Stoch-GFN in the figures) is robust to increasing trajectory lengths, and also performs well when the stochasticity level increases.
In addition, Stoch-GFN significantly outperforms strong baselines including MCMC, A2C, and SAC, discovering more modes faster.

\subsubsection{TF Bind 8 Generation} \label{sec:tfb}
\paragraph{Task.}
We now consider the practical task of generating DNA sequences with high binding activity with particular transcription factors, following~\citet{jain2022biological}.
At each time step, the agent appends a symbol from the vocabulary to the right of the current state.
As with the bit generation task, we consider a stochastic variant of the task following~\citet{yang2022dichotomy} with random actions taken with probability $\alpha$ (as described in Section~\ref{sec:grid_setup}).
We adopt a pre-trained neural network as the reward function following~\citet{jain2022biological} that estimates the binding activity.
We investigate how well Stochastic DB performs by comparing it with vanilla DB, MCMC, and RL-based methods including A2C and SAC.
For evaluation, we evaluate each method in terms of the number of modes with rewards above a threshold discovered in the batch of generated sequences.
We also use the mean reward and 50th percentile score for the top $100$ sequences ranked by their rewards from a batch of $2048$ generated sequences for each method as in~\citep{jain2022biological,trabucco2022design}.
We run each algorithm for $3$ different seeds, and report their mean and standard deviation.
We follow the same hyperparameters, architectures, and setup as in~\citet{jain2022biological}, and a detailed description of the setup can be found in Appendix A.3.

\begin{figure}[!h]
\centering
\subfloat[The number of modes.]{\includegraphics[width=0.8\linewidth]{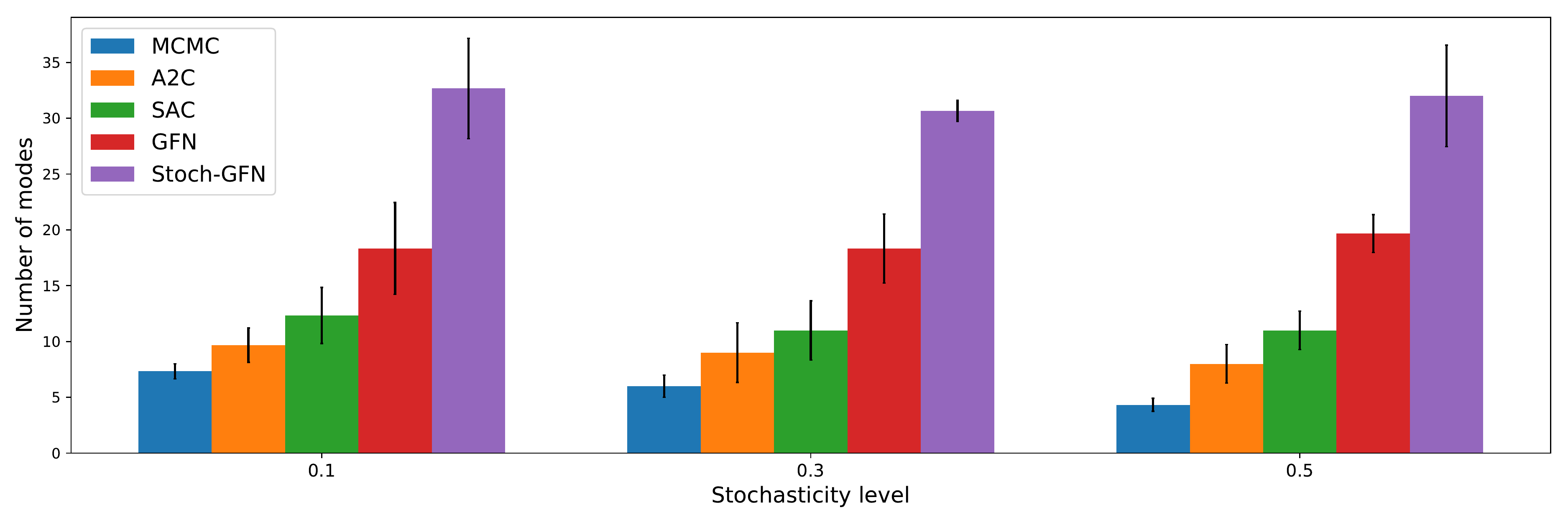}} \\
\subfloat[Top-100 reward (mean).]{\includegraphics[width=0.8\linewidth]{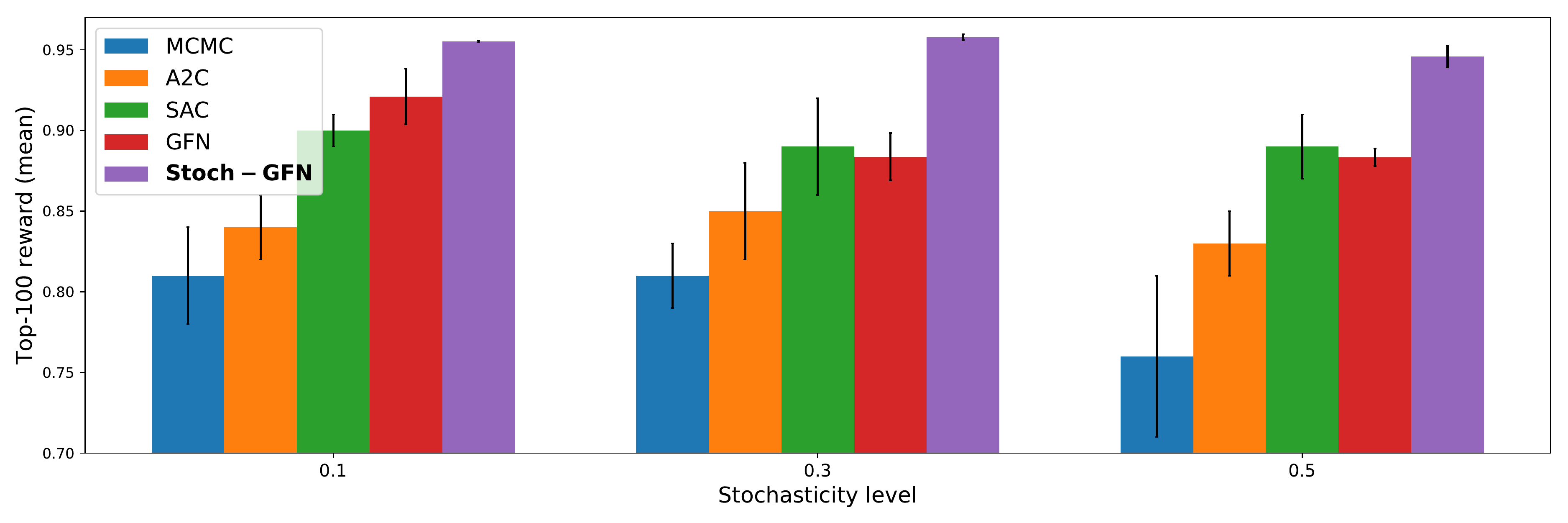}} \\
\subfloat[Top-100 reward (median).]{\includegraphics[width=0.8\linewidth]{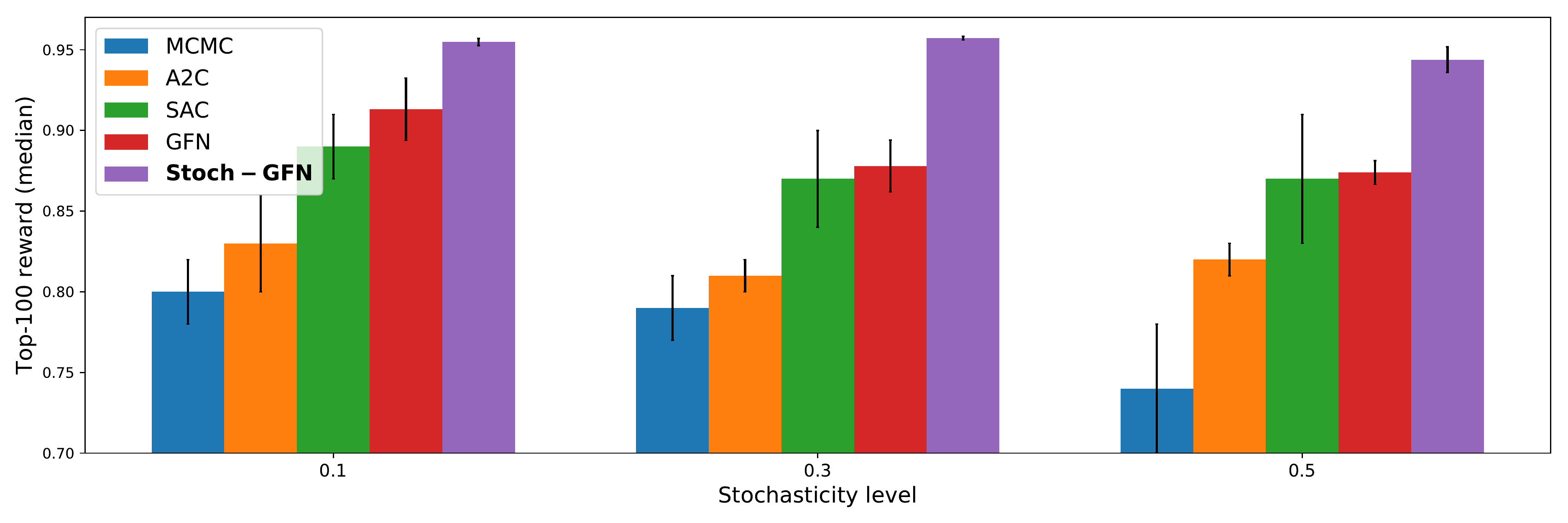}}
\caption{Results on the TF Bind 8 generation task, with better results for Stoch-GFN against MCMC, A2C, SAC and GFN baselines.}
\label{fig:tfb}
\end{figure}

\paragraph{Results.}
Comparison results of Stoch-GFN and baselines with varying stochasticity levels (ranging from $0.1$ to $0.5$) in terms of the number of modes discovered with rewards above a threshold during the training process and top-$100$ mean rewards are summarized in Figure~\ref{fig:tfb}.
As shown in Figure~\ref{fig:tfb}(a), Stoch-GFN discovers many more modes than GFN, MCMC, and RL-based methods in different stochasticity levels.
Stoch-GFN also achieves higher top-$100$ rewards (in mean and median) than baselines as demonstrated in Figures~\ref{fig:tfb}(b)-(c), where the top-$100$ reward of GFN decrease with an increasing stochastic level.
These results validate the effectiveness of Stoch-GFN in the more realistic task for biological sequence design with stochasticity in the environment.

\subsubsection{Antimicrobial Peptide Generation} \label{sec:amp}
\paragraph{Task.}
In this section, we study the realistic task of generating peptide sequences with anti-microbial properties~\citep{malkin2022trajectory,jain2022biological}. 
The agent chooses a symbol from the vocabulary that consists of $20$ amino acids and a special end-of-sequence action to the current state in a left-to-right manner at each time step.
The maximum length of the sequence is $60$, and the size of the resulting state space is $21^{60}$.
We consider a stochastic variant of the task (as in Section~\ref{sec:grid_setup}) with a stochasticity level of $\alpha = 0.1$.
The reward function is a pre-trained neural network that estimates the anti-microbial activity following~\citep{malkin2022trajectory} from the DBAASP database~\citep{pirtskhalava2021dbaasp}.
As in Section~\ref{sec:tfb}, we generate $2048$ sequences from each method and evaluate them in terms of the top-$100$ rewards and the number of modes discovered above a threshold.
We study the performance of Stochastic DB by comparing it with DB, MCMC, and RL-based methods.
We report the mean and standard deviation over $3$ runs for each method.
A detailed description of the setup is in Appendix A.4 following~\citet{malkin2022trajectory}.

\paragraph{Results.}
As shown in Table~\ref{tab:amp}, we observe that Stoch-GFN significantly outperforms GFN and other baselines in terms of the top-$100$ reward. In addition, it also discovers more modes with rewards above a threshold than baseline methods, which further validates its effectiveness on the more complex and challenging task.

\begin{table}[!h]
\caption{Better results with Stoch-GFN on the AMP generation task. Larger is better.}
\label{tab:amp}
\centering
\begin{tabular}{ccccc}
\toprule
 ~ & Top-$100$ reward & Number of modes \\
\midrule
MCMC  & $0.632\pm 0.035$& $3.67\pm 0.58$\\
A2C & $0.682\pm 0.032$ & $2.66\pm 0.58$\\
SAC & $0.754 \pm 0.047$& $4.33\pm 1.33$\\
GFN & $0.748 \pm 0.048$ & $3.0 \pm 3.0$\\
Stoch-GFN & {\highlight{$0.834 \pm 0.023$}} & {\highlight{$19.5 \pm 2.5$}}\\
\bottomrule
\end{tabular}
\end{table}

\section{Related Work}

\paragraph{GFlowNets.}
The universality and effectiveness of GFlowNets have been demonstrated in various kinds of applications, including biological sequence design~\citep{jain2022biological}, causal discovery and structure learning~\citep{deleu2022bayesian,nishikawa2022bayesian}, substructure learning of deep neural network weights via Dropout~\citep{Liu2022GFlowOutDW}, multi-objective optimization~\citep{jain2022multi}, and robust job scheduling problems~\citep{robust-scheduling}.
\citet{malkin2022trajectory} proposed the trajectory balance (TB) objective to optimize GFlowNet at a trajectory level instead of at the transition level as in detailed balance~\citet{bengio2021foundations}, but can induce large variance, where the problem is exacerbated in stochastic environments.
\citet{madan2022learning} propose the sub-trajectory balance method considers sub-trajectories.
The early GFlowNet proposals from \citet{bengio2021flow,bengio2021foundations} first formulated GFlowNets and pointed out possible future development directions. Originating from reinforcement learning, GFlowNets face the same long-term credit assignment challenges to propagate downstream reward signals to earlier states. 
\citet{Pan2023BetterTO} proposed a forward-looking GFlowNet formulation to exploit intermediate energies or rewards for more efficient credit assignment, making it possible to learn from incomplete trajectories.
\citet{pan2022gafn} incorporates intrinsic intermediate rewards into GFlowNets by augmenting the flow values for better exploration.
EB-GFN~\citep{zhang2022generative} jointly learns from data an energy/reward function along with the corresponding GFlowNet. \citet{zhang2022unifying} recently points out that the relationship between generative models and GFlowNets.
It is worth mentioning that \citet{zhang2023distributional} shares a similar goal to our work; it extends the GFlowNet framework for stochastic reward settings with distributional modeling, while this work focuses on stochasticity in the environment transition dynamics.

\paragraph{Model-based Reinforcement Learning.}
Model-based reinforcement learning (RL) is a promising approach for improved sample efficiency compared with model-free (RL) methods~\citep{lillicrap2015continuous,fujimoto2018addressing}, and has been successfully applied to many tasks such as robotics leveraging different dynamics models.
The stochastic value gradient method~\citep{heess2015learning} learns a hybrid of model-based and model-free RL which can learn stochastic policies in stochastic continuous control tasks.
Dreamer~\citep{hafner2019dream} learns latent dynamics to solve long-horizon tasks from high-dimensional images.
MuZero~\citep{antonoglou2021planning} combines model-based methods with Monte-Carlo tree search for planning, and it has achieved great success in game playing. Stochastic MuZero~\citep{schrittwieser2020mastering} learns a stochastic model for extending MuZero to stochastic environments. 

\section{Conclusion}
In this paper, we introduce a new methodology, Stochastic GFlowNets, which is the first empirically effective approach to extend GFlowNets to the more general and realistic stochastic environments, where existing GFlowNet methods can fail.
Our method learns the GFlowNet policy and also the environment model to capture the stochasticity in the environment.
We conduct extensive experiments in standard tasks for benchmarking GFlowNets with stochastic transition dynamics. Results show that Stochastic GFlowNet learns significantly better than previous methods in the presence of stochastic transitions. 
It is interesting for future work to study advanced model-based approaches for approximating the transition dynamics, and also apply our method to other challenging real-world tasks.

\section*{Acknowledgements}
The authors would like to thank Almer Van der Sloot, Kanika Madan, and Qingpeng Cai for insightful discussions about the paper and the baselines in the AMP generation task.
Longbo Huang is supported in part by the Technology and Innovation Major Project of the Ministry of Science and Technology of China under Grant 2020AAA0108400 and 2020AAA0108403, and Tsinghua Precision Medicine Foundation 10001020109. 
Yoshua Bengio acknowledges the funding from CIFAR, Genentech, Samsung, and IBM. 

\bibliography{pan_56}

\clearpage
\appendix

\section{Experimental Details}

\subsection{Gridworld} \label{app:grid}
The reward function for GridWorld is defined as in Eq.~\eqref{eq:grid_r} following \citet{bengio2021flow}, where $R_0=2.0$, $R_1=0.5$, and $R_2=0.001$.
\begin{equation}
\begin{split}
R(x)&=R_0+R_1 \prod_i \mathbb{I}\left(0.25<\left|x_i / H-0.5\right|\right)\\
&+R_2 \prod_i \mathbb{I}\left(0.3<\left|x_i / H-0.5\right|<0.4\right)
\end{split}
\label{eq:grid_r}
\end{equation}
We use a feedforward network that consists of two hidden layers with $256$ hidden units and LeakyReLU activation. States are represented using one-hot embeddings. As for the environment model in Stochastic GFlowNet, it is also a feedforward layer consisting of two hidden layers with $256$ hidden units and LeakyReLU activation.
All models are trained for $20000$ iterations, and we use a parallel of $16$ rollouts in the environment at each iteration (which are then stored in the experience replay buffer). The GFlowNet model is updated based on the rollouts, and we train it based on the Adam~\citep{kingma2014adam} optimizer using a learning rate of $0.001$ (the learning rate for $Z$ in TB is $0.1$).
We train the environment model using data sampled from the experience replay buffer with a batch size of $16$, which is trained using the Adam optimizer with a learning rate of $0.0001$.
MCMC and PPO use the same configuration as in \citet{bengio2021flow}.

\subsection{Bit Sequences} \label{app:bit}
We follow the same setup for the bit sequence generation task as in \citet{malkin2022trajectory}. The GFlowNet model is a Transformer~\citep{vaswani2017attention} that consists of $3$ hidden layers with $64$ hidden units and uses $8$ attention heads. 
The exploration strategy is $\epsilon$-greedy with $\epsilon=0.0005$, while the sampling temperature is set to $1$. 
It uses a reward exponent of $3$.
The learning rate for training the GFlowNet model is $5 \times 10^{-3}$, with a batch size of $16$.
As for the environment model in Stochastic GFlowNet, we use a feedforward network consisting of two hidden layers with $2048$ hidden units and ReLU activation, which is trained using the Adam optimizer with a learning rate of $5 \times 10^{-4}$.
It is trained using data sampled from the experience replay buffer with a batch size of $128$.
We train all models for $50000$ iterations, using a parallel of $16$ rollouts in the environment.
MCMC, A2C, and SAC adopt the same configuration as in \citet{malkin2022trajectory}.

\subsection{TFBind-8} \label{app:tfb}
For the TFBind-8 generation task, we follow the same setup as in \citet{jain2022biological}.
The vocabulary consists of $4$ nucleobases, and the trajectory length is $8$.
The GFlowNet model is a feedforward network that consists of $2$ hidden layers with $2048$ hidden units and ReLU activation.
The exploration strategy is $\epsilon$-greedy with $\epsilon=0.001$, while the reward exponent is $3$.
The learning rate for training the GFlowNet model is $10^{-4}$, with a batch size of $32$.
As for the environment model, we use a feedforward network consisting of two hidden layers with $2048$ hidden units and ReLU activation, which is trained using the Adam optimizer with a learning rate of $10^{-5}$.
It is trained using data sampled from the experience replay buffer with a batch size of $16$.
We train all models for $5000$ iterations.
MCMC, A2C, and SAC baselines follow the same configuration as in \citet{jain2022biological}.

\subsection{Antimicrobial Peptide Generation} \label{app:amp}
We follow the same setup for the antimicrobial peptide generation task as in \citet{malkin2022trajectory}.
The GFlowNet model is a Transformer~\citep{vaswani2017attention} that consists of $3$ hidden layers with $64$ hidden units and uses $8$ attention heads. 
The exploration strategy is $\epsilon$-greedy with $\epsilon=0.01$, while the sampling temperature is set to $1$. 
It uses a reward exponent of $3$.
The learning rate for training the GFlowNet model is $0.001$, with a batch size of $16$.
As for the environment model, we use a feedforward network consisting of two hidden layers with $128$ hidden units and ReLU activation, which is trained using the Adam optimizer with a learning rate of $0.0005$.
It is trained using data sampled from the experience replay buffer with a batch size of $128$.
We train all models for $20000$ iterations, using a parallel of $16$ rollouts in the environment.

\end{document}